\documentclass[twoside]{article}

\usepackage[accepted]{aistats2024}
\usepackage{natbib}
\setcitestyle{authoryear} 
\usepackage{algorithm}
\usepackage{mathtools}
\usepackage{amssymb}
\usepackage{amsmath, amsthm}
\usepackage{algpseudocode}
\usepackage{graphicx}
\usepackage{svg}
\usepackage{microtype}
\usepackage{graphicx}
\usepackage{subfigure}
\usepackage{multirow}
\usepackage{hyperref}

\newtheorem{theorem}{Theorem}

\theoremstyle{definition}

\theoremstyle{remark}

\begin{document}

%

%

\twocolumn[

\aistatstitle{GRAWA: Gradient-based Weighted Averaging for Distributed Training of Deep Learning Models}

\aistatsauthor{ Tolga Dimlioglu \And Anna Choromanska }

\aistatsaddress{ New York University  \And  New York University} ]

\begin{abstract}
We study distributed training of deep learning models in time-constrained environments. We propose a new algorithm that periodically pulls workers towards the center variable computed as a weighted average of workers, where the weights are inversely proportional to the gradient norms of the workers such that recovering the flat regions in the optimization landscape is prioritized. We develop two asynchronous variants of the proposed algorithm that we call Model-level and Layer-level Gradient-based Weighted Averaging (resp. MGRAWA and LGRAWA), which differ in terms of the weighting scheme that is either done with respect to the entire model or is applied layer-wise. On the theoretical front, we prove the convergence guarantee for the proposed approach in both convex and non-convex settings. We then experimentally demonstrate that our algorithms outperform the competitor methods by achieving faster convergence and recovering better quality and flatter local optima. We also carry out an ablation study to analyze the scalability of the proposed algorithms in more crowded distributed training environments. Finally, we report that our approach requires less frequent communication and fewer distributed updates compared to the state-of-the-art baselines. 
\end{abstract}

\vspace{-3mm}
\section{Introduction}
Training deep learning models in a distributed setting has become a necessity in scenarios involving large models and data sets due to efficiency and practicality. There are different ways of parallelizing the computations in distributed environments, such as data and model parallelization, both of which has its own advantages and use cases \citep{demyst_dist}. These parallelization techniques were used in many different machine learning tasks such as image recognition \citep{imagenet}, machine translation \citep{machine_translation}, language understanding \citep{bert_nlp}, and more. This paper focuses on data parallelization in which models on different devices are trained with different portions from the dataset. \\

In data parallelization schemes, many models are simultaneously trained on different devices with different and non-overlapping portions of the data set, i.e., the data portions on different devices are mutually exclusive \citep{demyst_dist}.In this way, the effective number of data samples used for training is increased by the number of devices in the parallel computing environment. When training deep learning models, most of the time these devices are Graphical Processing Units (GPUs) or Neural Processing Units (NPUs) \citep{dist_survey}. During training, each worker (GPU) on each node runs Stochastic Gradient Descent (SGD) algorithm~\citep{bottou-98x} or its variant \citep{kingma2014adam, overviewSGD}. The communication between workers and the distributed updates of model parameters done across the nodes assure that the knowledge embedded in the model parameters on different devices is properly shared among all workers. \\

In this paper we introduce two new algorithms for distributed deep learning optimization in data parallel setting. We refer to them as Gradient-based Weighted Averaging (GRAWA) algorithms, Model-level GRAWA (MGRAWA) and Layer-level GRAWA (LGRAWA). By design, both of these variants address the curse of symmetry problem~\citep{lsgd_paper} which occurs when the model is stuck on the hill in the optimization landscape lying in-between workers that are trapped in the local minima around it, and simultaneously prevent workers from getting stuck in narrow minima by pushing them towards flatter regions on the optimization loss surface. The proposed methods compare favorably to SOTA approaches in terms of convergence speed, generalization ability, and communication overhead, by considering the geometry of the landscape. Although there exist flat minima seeking deep learning optimizers in the literature, such as Entropy-SGD \citep{eSGD}, SAM \citep{SAM}, and LPF-SGD \citep{lpf-sgd}, to the best of our knowledge, a flatness-aware update policy has never been introduced before in the context of distributed deep model training, more specifically for parameter-sharing methods. Our approach is therefore new. Our MGRAWA algorithm encourages flat minima by considering the gradient scores of the workers in the distributed training environment. Finally, the LGRAWA algorithm that we propose prioritizes model layers appropriately, when computing worker averages and model updates, to take advantage of the layers that learn faster than the others. This is motivated by the fact that neural networks are essentially compositions of functions and we seek a robust solution for each function in the composition. In this sense, our treatment of model layers and their effect on the algorithm's distributed update is also novel. \\

This paper is organized as follows: Section~\ref{sec:RW} reviews the related literature, Section~\ref{sec:ProblemFormulation} formulates the problem, Section~\ref{sec:GRAWA} motivates, derives the GRAWA algorithm and provides theoretical analysis, Section \ref{sec:GRAWAvariants} introduces two GRAWA variants (MGRAWA and LGRAWA), Section~\ref{sec:Experiments} presents the experimental results and ablation study on scalability, and Section~\ref{sec:Conclusion} concludes the paper. The supplement contains experiment details, additional results and proofs.\\

\vspace{-3mm}
\section{Related Work}
\label{sec:RW}

In this section we review the literature devoted to distributed data parallel deep learning optimization and methods encouraging the recovery of flat minima in a deep learning optimization landscape. \\

\textbf{Distributed data parallel deep learning optimization} There exist only a few algorithms for distributed data parallel deep learning optimization. The DataParallel algorithm~\citep{data_parallel}, a gradient-sharing method, initiates all of the workers on different devices with the same model parameters that also remain the same throughout the training. The communication between the workers occurs after all the nodes complete processing their data batches. During the communication, the calculated gradients on different nodes are averaged and used to update the model parameters. Since communication occurs after processing each batch, the method suffers from communication overhead. Furthermore, since the models are kept identical on all the machines throughout the training, the algorithm does not take advantage of having multiple workers performing individual exploration of the loss surface \citep{loss_surface}. \\


Moving on to the parameter sharing methods, in the EASGD algorithm \citep{EASGD}, each worker has distinctive model parameters during the course of training since they all run SGD independently on different data shards. During the communication, the algorithm applies an elastic force to each worker pulling it towards the center model. Here, the center model is calculated as a moving average of the model parameters of all the workers, both in space and time. The elastic force, therefore, establishes the link between different workers. The method suffers from the curse of symmetry phenomenon that occurs when the center model gets stuck at the maximum between several local minima that trap the workers. This can happen in the in the symmetric regions of the loss landscape. The workers are then pushed towards the maximum point during the distributed training phase which impedes their generalization abilities. The LSGD algorithm~\citep{lsgd_paper} addresses this problem by pushing all of the workers towards the leader worker, which is selected as the worker with the smallest training loss. Despite breaking the curse of symmetry and outperforming both DataParallel and EASGD in terms of convergence speed and accuracy, the LSGD algorithm can recover sub-optimal final model when the leader worker is trapped in a narrow region of the loss landscape, as we will show in the motivating example later in the paper. \\

\textbf{Flatness-aware deep learning optimization} It has been shown in the literature that model's generalization capability in deep learning is firmly tied to the flatness of the loss valley that the model converged to \citep{fantastic_measures}. In particular, when the model converges to flatter local minima, it yields a smaller generalization gap. There are many flatness measures introduced in the literature such as $\epsilon$-sharpness \citep{lb_sharpness}, Shannon entropy of the output layer \citep{shannon_entropy}, PAC-Bayes measures \citep{fantastic_measures} and, measures based on the Hessian and its spectrum \citep{fantastic_measures, hessian1, hessian2}. However, they all are expensive to compute in a time-constrained environment. \\

There are also optimizers with mechanisms for seeking good quality \citep{poor_local_minima}, flat-minima. Entropy-SGD \citep{eSGD} smooths out the energy landscape of the original loss function using local entropy, while increasing the processing time of a batch by $L$ folds, where $L$ is the number of Langevin iterations. The SAM optimizer \citep{SAM}  seeks parameters that lie in neighborhoods having uniformly low loss. Their optimization problem is formulated as a min-max problem. They approximate the inner maximization problem in order to derive a closed-form solution, whose computation requires double iteration for one batch. In LPF-SGD \citep{lpf-sgd}, the gradient is smoothed by taking the average of the gradient vectors calculated after applying Gaussian noise on the current batch. This process multiplies the single batch processing time by the number of Markov Chain Monte-Carlo (MCMC) sampling iterations. \\

Notice that it is unclear how to adapt these schemes efficiently to the distributed setting since it would require distributing the computations of flatness measure across multiple workers. Also, since the current flat minima seeking optimizers require extra time to compute flatness measures, they are not well suited for data parallel training, where the goal is to reduce the training time. A flat minima seeking update has to be simple in order to avoid introducing computation overhead and is easy to parallelize. This is challenging and has never been investigated before to the best of our knowledge.

\section{Problem Formulation}
\label{sec:ProblemFormulation}

Consider the problem of minimizing a loss function $F$ with respect to model parameters $x$ over a large data set $\mathcal{D}$. In a parallel computing environment with $M$ workers $(x_1, x_2, ..., x_M)$, this optimization problem can be written as:

\vspace{-7mm}
\begin{equation}
  \begin{aligned}
    \min_{x} &\quad F(x; \mathcal{D}) = \\ 
    &\min_{x_1,\dots, x_M} \sum_{m=1}^{M} \mathbb{E}_{\xi \sim \mathcal{D}_m } f (x_m; \xi) + \frac{\lambda}{2} || x_m - x_c ||^2
  \end{aligned}
  \vspace{-3mm}
  \label{eq:dist_opt}
\end{equation}

where $\mathcal{D}$ is partitioned and sent to $m$ workers and, $\mathcal{D}_m$ is the sampling distribution of worker $m$ exclusive to itself. $x_c$ is the center variable (for example, an average of all the workers), and $x_m$ stands for the parameters of model $m$. The equivalence of the optimization problems captured on the left-hand side and right-hand side of the Equation \ref{eq:dist_opt} is investigated in the literature and is commonly referred to as the global variable consensus optimization problem \citep{concensus_opt}. The penalty term $\lambda$ in front of the quadratic term in Equation \ref{eq:dist_opt} ensures the proximity of the model parameters of different workers by attracting them towards the center model. 

\vspace{-3mm}
\section{GRAWA Algorithm}
\label{sec:GRAWA}
\vspace{-3mm}
\subsection{Motivating Example}
\label{motivation}

Motivated by the connection between the generalization capability of deep learning models and the flatness of the local minima recovered by deep learning optimizers, we propose an algorithmic family called GRAWA that pushes the workers to flatter regions of the loss surface in the distributed optimization phase. Before formalizing the algorithm, we first simply illustrate its effectiveness in seeking flatter valleys on toy non-convex surface. Let us consider the Vincent function that has the following formulation for a $2$-D input vector: $f(x,y) = - \sin \left ( 10\ln (x) \right ) - \sin \left ( 10\ln (y) \right ).$ 

\vspace{-4mm}
\begin{figure}[H]
\centering
\includegraphics[width=0.9\linewidth]{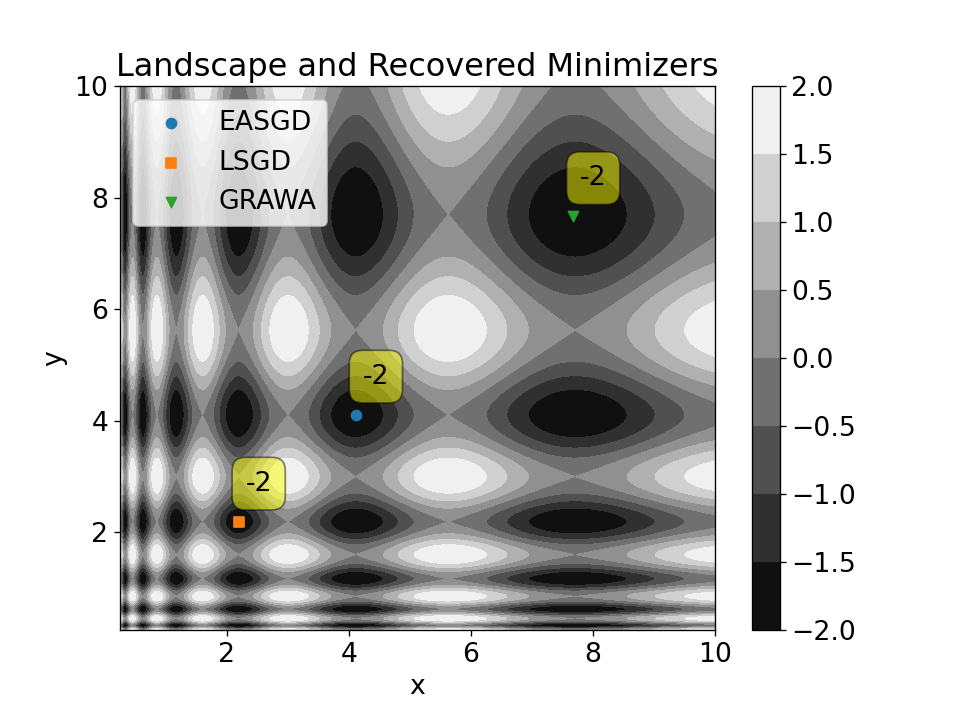}
\vspace{-5mm}
\caption{Contour plot of the loss landscape of the Vincent function and the final optima obtained by running LSGD, EASGD, and GRAWA optimizers.} 
\centering
\label{fig:converged_points}
\end{figure}
\vspace{-3mm}
The Vincent function is a non-convex function and it has several global minimizers that all attain a value of $-2$. But the flatness of the valleys in which the minimizers reside is different from each other. The contour map of the loss surface and the final solutions obtained by running the EASGD, LSGD, and GRAWA optimizers are provided in the figure below. As can be seen from Figure \ref{fig:converged_points}, all algorithms converge to a point that corresponds to the same loss value, but GRAWA recovers the valley in the loss landscape that is flatter than in the case of the other two schemes. The convergence trajectories of different distributed optimizers can be found in Appendix \ref{appendix:motivational_example}.

\subsection{Gradient Norm and Flatness}
Our motivation behind finding a mechanism that encourages the recovery of flat loss valleys relies on simple properties of the gradient vector. If the Euclidean norm of the gradient vector is high, it implies that the point at which the gradient is calculated is still at the steep slope of the loss valley. Otherwise, if the Euclidean norm of the gradient vector is low, the point at which the gradient is calculated is in the flat region of the loss landscape. As mentioned in the related work, flatness and generalization are mirror terms. Therefore, for completeness, we confirm empirically that for the workers residing in different valleys, the valleys with worse generalization capabilities would yield greater gradient norms, i.e., they would be sharper. We train $162$ ResNet-18 models with different hyperparameters on the CIFAR-10 dataset in a parallel computing environment using the EASGD distributed training method. For each trained model, we calculate the norm of the gradient vector accumulated on the whole train set, the train error, and the test error. We indeed find out that there is a relation between the generalization gap (i.e., the difference between the test and train errors) and the gradient norm (and thus the flatness). 

\vspace{-5mm}
\begin{figure}[h!]
\centering
\includegraphics[width=0.85\linewidth]{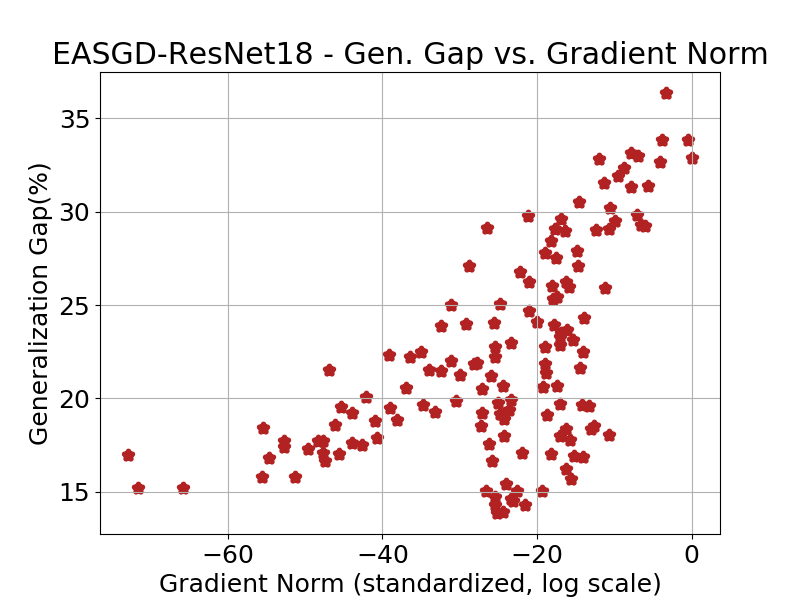}
\caption{Plot of standardized gradient norms in log scale vs. the generalization gap $(\%)$.} 
\label{fig:grad_norm}
\vspace{-1mm}
\end{figure}

As can be seen in Figure \ref{fig:grad_norm}, when the generalization gap increases (the model's generalization capability becomes worse), the gradient norm increases as well (the flatness deteriorates). Hence, we can use the gradient norm to obtain flatter valleys with good generalization capabilities. In a distributed setting, we propose to apply a weighted averaging scheme that favors workers with small gradient norms and penalizes the ones with large gradient norms when calculating the center model. Using gradient norm as opposed to common flatness metrics, such as $\epsilon-$sharpness \citep{lb_sharpness}, Hessian-based measures \citep{hessian}, or recently introduced LPF measure \citep{lpf-sgd}, is essential since these metrics are computationally expensive to periodically calculate in a time-constrained distributed training environment.

\vspace{-2mm}
\subsection{GRAWA: Gradient-based Weighted Averaging}

We next explain the GRAWA algorithm, from which MGRAWA and LGRAWA schemes originate.

\subsubsection{GRAWA Method}
Gradient-based Weighted Averaging (GRAWA) algorithm is an asynchronous distributed training algorithm that periodically applies a pulling force to all workers towards a consensus model created by taking weighted average among the workers. The weights in this weighted averaging scheme are determined based on the gradients. The vanilla GRAWA algorithm assigns to each worker $m$ the weight $\beta_m$ that is inversely proportional to the Euclidean norm of the flattened gradient vector $A_m$. More specifically, the weights are calculated as follows:

\vspace{-6mm}
\begin{equation}
\begin{aligned}
\beta_m \propto \frac{1}{||A_m||_2}, \hspace{0.25em} &\sum_{m=1}^{M} \beta_{m} = 1 \hspace{0.25em} \longrightarrow \hspace{0.25em} \beta_m = \frac{\Theta}{||\nabla f(x_m)||} \\ 
\text{where} \hspace{0.5em} \Theta &= \frac{ \prod_{i=1}^M ||\nabla f(x_i)||}{  \sum_{i=1}^M \frac{\prod_{j=1}^M ||\nabla f(x_j)||}{||\nabla f(x_i)||} } \\
\end{aligned}
\label{eq:grawa}
\end{equation}

Let $x_m$ be the model parameters of the worker $m$. Then, the center model $x_C$, is calculated with a weighted average. Thus, in each distributed training phase, a new $x_C$ is calculated and then all the workers are pushed towards it with a fixed $\lambda$ coefficient where $\lambda$ is $\in [0,1]$. This coefficient is the strength of the pulling force applied to the workers. Overall, the calculation of the center variable and the distributed update are as follows:

\vspace{-6mm}
\begin{equation*}
x_C = \sum_{m=1}^{M} \beta_{m} x_m \quad \quad \text{and} \quad \quad x_m \leftarrow (1-\lambda) x_m + \lambda x_C.
\end{equation*}

\subsubsection{Convergence Rate of GRAWA in Convex Case}
We show that GRAWA algorithm matches the theoretical guarantees of both EASGD and LSGD and enjoys the same convergence rate as SGD. We first prove that center variable $(x_C)$ obtained with GRAWA algorithm attains a smaller loss value than the workers in the distributed training environment. We also note that the convergence analysis for MGRAWA would be similar to GRAWA with modifications to the constants in the derivation whereas, for LGRAWA, such analysis would be mathematically intractable since the weighted averaging is done on the layer level.

\begin{theorem}
\label{main_body:smaller_th}
Let $x_C = \sum_{i=1}^{M} \beta_i x_i$ and $\beta_i$'s are calculated as in equation \ref{prelim:grad_weights}. For an $L$-Lipschitz differentiable, real-valued, continuous convex function $f$ with minimizer $x^*$ that also satisfies $||\nabla f(x) || \geq \mu \left (f(x) -f(x^*) \right)$ and is also bounded by a cone which has a slope $k$ and a tip at $x^*$; the GRAWA center variable holds the following property: $f(x_C) \leq f(x_i)$ for all $i \in 1,2,...,M$ when $k \mu \geq L \sqrt{M}$. 
\end{theorem}

We consider a general update rule where the distributed update is applied at every iteration. Thus, the overall update rule for each iteration becomes $x^{t+1}_i = x^t_i - \eta \left (  \nabla f(x^t_i) + \lambda(x^t_i - x_C) \right )$. Note that this update rule is applicable to each worker $i$. For simplicity, we drop the worker indices for the following theorem that is derived from \citep{lsgd_paper}.

\begin{theorem}
Let $f$ be a function that satisfies the conditions of \ref{main_body:smaller_th}. Let $\tilde{g}(x)$ be an unbiased estimator for $\nabla f(x)$ with $\mathrm{Var}(\tilde{g}(x)) \leq \sigma^2 + \nu ||\nabla f(x)||^2$, and let $x_C$ be the center variable obtained with the GRAWA algorithm. Suppose that $\eta, \lambda$ satisfy $\eta \leq \left ( 2L(\nu + 1)  \right)^{-1}$ and $\eta \lambda \leq \frac{m}{2L}$, $\eta \sqrt{\lambda} \leq \frac{\sqrt{m} }{\sqrt{2}L}$. Then the GRAWA step satisfies:
\vspace{-1mm}
\begin{align*}
    \mathbb{E}[f(x^{t+1})] - f(x^*) \leq &(1-m\eta)(f(x^t) - f(x^*)) \\
    - &\eta \lambda (f(x^t)-f(x_C)) + \frac{\eta^2 L}{2} \sigma^2
\end{align*}

\vspace{-1mm}
The presence of the new term $f(x_C)$ due to the GRAWA update rule increases the speed of the convergence since $f(x_C) \leq f(x)$ as given in \ref{main_body:smaller_th}. Then, $\limsup_{t \rightarrow \infty} \mathbb{E}[f(x^{t+1})] - f(x^*) \leq \eta \frac{L}{2m} \sigma^2 $. If $\eta$ decreases at rate $\eta = O(\frac{1}{t})$, then $\mathbb{E}[f(x^{t+1})] - f(x^*) \leq O(\frac{1}{t})$.

\end{theorem}

\subsubsection{Convergence Rate of GRAWA in Non-Convex Case}

Here we state the convergence guarantee for GRAWA in the non-convex optimization setting. 

\begin{theorem}
Let $F$ be an $L$-smooth function and let us have the following bounds for the stochastic gradients that satisfy the following variance bounds: $\mathbb{E}_{\xi \sim \mathcal{D}_m} || \nabla f(x, \xi) - \nabla f_m(x_m) ||^2 \leq \sigma^2$, $\mathbb{E}_{m \sim U[M]} || \nabla f_m(x_m) - \nabla F(x_m) ||^2 \leq \zeta^2$. Also, assume that we have the following bound for local workers and center variable: $\mathbb{E} || x_m^t - x_c^t  ||^2 \leq \rho^2$ at any time or iteration $t$. We show that GRAWA algorithm and its variants satisfy the following convergence rate:  
\vspace{-1mm}
\begin{equation*}
\begin{aligned}
\frac{1}{MN}&\sum_{m=1}^{M} \sum_{t=1}^{N} \mathbb{E}|| \nabla F(x_m^t) ||^2 \leq  \\
& O \left ( \frac{F(x^0) - F(x^*)}{MN\eta}  + \left ( 2L\zeta^2 + L\sigma^2 + \lambda^2\rho^2L \right ) \eta \right ),
\end{aligned}
\end{equation*}
which characterizes the square norm of the gradients averaged with respect to all $M$ workers and the number of samples.

\end{theorem}




\section{GRAWA Algorithm Variants}
\label{sec:GRAWAvariants}

In this paper we also develop two extensions of the GRAWA algorithm which are more suitable for the training of deep learning models: Model-level GRAWA (MGRAWA) and Layer-level (LGRAWA). To be more specific, the mechanism of GRAWA that relies on simply vectorizing all the gradients and then taking their norm is less representative than individually calculating the layer gradient norms and summing them due to the triangle inequality. Furthermore, in these variants, we are utilizing the stochastic gradients for norm calculation. The details of MGRAWA and LGRAWA are provided in the following subsections.

\subsection{MGRAWA Algorithm}

In the MGRAWA scheme, similar to the GRAWA method, the center model is obtained by taking the weighted average of the model parameters of all the workers. First, before any updates are applied, the gradients are accumulated in each worker using a randomly selected subset from the train set. Unlike the vanilla GRAWA, these accumulated gradients in all of the layers are summed up for each worker. This quantification better represents the gradient norm of each layer individually. Let $g^n_{k}$ be the stochastic gradient calculated in the $k^{th}$ layer of the model with respect to the $n^{th}$ sample. Also, let $N$\footnote{In practice, we choose N equal to the batch size.} be the size of the subset of the train set, and $K$ be the number of layers in the network. Then, the accumulated gradient $A_m$ for each worker $m$ is calculated as follows:

\vspace{-5mm}
\begin{equation}
G_{k} = \sum_{n}^{N} g^n_{k} \quad \quad \quad \quad A_m = \sum_{k}^{K} ||G_{k}||_{F},
\label{eq:model_level_acc}
\vspace{-4mm}
\end{equation}

where $||\cdot||_{F}$ is the Frobenius norm. Let us have $M$ workers in the distributed training and let $A_1, A_2, ... A_M$ be the sum of accumulated gradients of all layers for worker $1$ to worker $M$ respectively. Then, the weight calculated for worker m, $\beta_m$, is inversely proportional to its accumulated gradient $A_m$. Furthermore, we require that the sum of these weights equals $1$.
\vspace{-1mm}
\begin{equation}
\beta_m \propto \frac{1}{A_m} \quad \quad \text{and} \quad \quad \sum_{m=1}^{M} \beta_{m} = 1
\label{eq:weights_model}
\vspace{-3mm}
\end{equation}

The expression to calculate the center model is the same as in GRAWA. Pseudo-code of the MGRAWA algorithm is provided in the Algorithm \ref{algorithm:MGRAWA_short} (proximity search step from the pseudo-code will be explained later).

\begin{algorithm}[ht]
    \caption{MGRAWA}
    \label{algorithm:MGRAWA_short}
    \begin{algorithmic}
        \State \textbf{Input}: Pulling force $\lambda$, communication period $\tau$, learning rate $\eta$, proximity search strength $\mu$, loss function $f$
        \State \textbf{Initialize} workers $x_1, x_2, ..., x_M, x_C$ from the same random model, worker-exclusive data shards $\Psi_1, \Psi_2, ..., \Psi_M$ and iteration counters for workers $t_1=t_2= ... =t_M=0$ 
        \State At each worker $m$ \textbf{do} 
        \While{not converged}
        \State Draw a random batch $\xi_m \in \Psi_m$
        \State $x_{m} \leftarrow x_{m} - \eta \nabla f(x_m; \xi_m) $  
        
        \State $x_m \leftarrow \left(1-\frac{\mu}{\tau} \right) x_m + \frac{\mu}{\tau} x_C$ (Prox.)
        \State $t_m \leftarrow t_m + 1$
        
        \If{$M\tau$ divides $\sum_{m=1}^{M} t_m$}
        \State Draw a random batch from $\mathcal{D}$
        \State  (this batch is same for all workers)
        \State Accumulate the gradients
        \State Calculate the weights $\beta_m$
        \State $x_C = \sum_{m=1}^{M} \beta_{m} x_m$   
        \State $x_m \leftarrow (1-\lambda) x_m + \lambda x_C$ 
        \EndIf
        \EndWhile
    \end{algorithmic}
\end{algorithm}

\subsection{LGRAWA Algorithm}

\begin{figure}[ht]
    \centering
    \includegraphics[width=0.75\columnwidth]{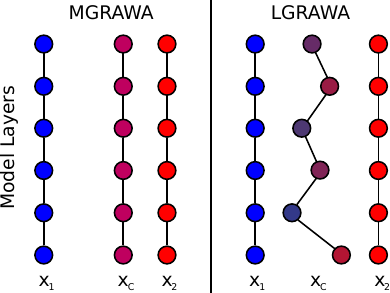}
    \caption{Illustration of MGRAWA and LGRAWA weighted averaging schemes in the case of two workers.}
    \label{fig:mgrawa_vs_lgrawa}
\end{figure}

In the LGRAWA algorithm, the weighted averaging scheme is also used, but it is applied independently on each layer of the network. Similarly to MGRAWA, first gradients are accumulated for the network layers using the subset of the train data set without updating any network components. The main difference in this version is that, rather than assigning a weight to the whole model parameter vector, weights do vary across different model layers. The motivation behind this approach relies on favoring a worker's layer that corresponds to a gradient with a smaller norm compared to the same layer of the other workers. Intuitively, a smaller gradient norm means that the layer requires less correction. We refer to these layers as \textit{mature} layers and, in LGRAWA's weighted averaging scheme, mature layers are granted larger weights. Figure~\ref{fig:mgrawa_vs_lgrawa} provides an illustration of the difference between MGRAWA and LGRAWA algorithms. In order to describe LGRAWA mathematically, let us define $A^k$ that corresponds to the norm of the total accumulated gradient at layer $k$. We can write $A^k$ as:

\vspace{-5mm}
\begin{equation}
G_{k} = \sum_{n}^{N} g^n_{k} \quad \quad \text{and} \quad \quad A^k = ||G_{k}||_{F}.
\label{eq:layer_level_acc}
\vspace{-4mm}
\end{equation}

Let $A^k_m$ be the norm of the accumulated gradient of worker $m$ at its $k^{th}$ layer. For each worker $m$, we form the following list:
\vspace{-1mm}
\begin{equation}
[A^k_m]_{k=1}^{K} = [A^1_m, A^2_m, ..., A^K_m].
\label{eq:layer_level_gradlist}
\vspace{-3mm}
\end{equation}

Let $\beta_m^k$ be the coefficient for worker $m$'s $k^{th}$ layer. Then, for each layer of each worker, we find the coefficients of the weighted average scheme using inverse proportionality to gradient norm scores. Notice that the coefficient calculation procedure is applied to each layer $k$ separately and $\sum_{k=1}^{K} \sum_{m=1}^{M} \beta_m^k = K$. Let $x^k_m$ and $x^k_C$ be the parameters at layer $k$ of worker $m$ and the center model respectively. Overall, we write:

\vspace{-6mm}
\begin{equation}
\beta_m^k \propto \frac{1}{A^k_m} \text{  s.t.  } \sum_{m=1}^{M} \beta^k_m = 1  \longrightarrow  x^k_C = \sum_{m=1}^{M} \beta_{m}^{k} x_m^k
\label{eq:layer_level_coeff}
\end{equation}

\vspace{-3mm}
Equation \ref{eq:layer_level_coeff} calculates the parameters of the consensus model at the $k^{th}$ layer in the network structure. So, the consensus model is simply the model that is constructed from layers $k = 1:K$, where the parametrization of each layer is calculated according to this equation. We can symbolically write $x_C = \text{construct}( x_C^1, x_C^2, ..., x_C^K)$. Then, the update rule is the same as in MGRAWA and GRAWA updates. Pseudo-code of can be found in Algorithm \ref{alg:LGRAWA}. \\

\begin{algorithm}[ht]
\caption{LGRAWA}
\label{alg:LGRAWA}
\begin{algorithmic}
\State \textbf{Input}: Pulling force $\lambda$, communication period $\tau$, learning rate $\eta$, proximity search strength $\mu$, loss function $f$
\State \textbf{Initialize} workers $x_1, x_2, ..., x_M, x_C$ from the same random model, worker-exclusive data shards $\Psi_1, \Psi_2, ..., \Psi_M$ and iteration counters for workers $t_1=t_2= ... =t_M=0$ 

\State At each worker $m$ \textbf{do} 
\While{not converged}
\State Draw a random batch $\xi_m \in \Psi_m$
\State $x_{m} \leftarrow x_{m} - \eta \nabla f(x_m; \xi_m) $  
\State $x_m \leftarrow \left(1-\frac{\mu}{\tau} \right) x_m + \frac{\mu}{\tau} x_c$ (Prox.)
\State $t_m \leftarrow t_m + 1$
\vspace{1mm}

\If{$M\tau$ divides $\sum_{m=1}^{M} t_m$}
\State Draw a random batch from $\mathcal{D}$ 
\State  (this batch is same for all workers)
\State Accumulate the gradients 
\State Obtain the list $[A^1_m, A^2_m, ..., A^K_m] $
\State Calculate $\beta_m^k$ for all $k=1:K$ 
\State Calculate $x^k_C = \sum_{m=1}^{M} \beta_{m}^{k} x_m^k$
\State $x_C = \text{construct}( x_C^1, x_C^2, ..., x_C^K)$    
\State $x_m \leftarrow (1-\lambda) x_m + \lambda x_C$ 
\EndIf
\EndWhile
\end{algorithmic}
\end{algorithm}

In Figure \ref{fig:mgrawa_vs_lgrawa}, we illustrate the difference between MGRAWA and LGRAWA algorithms. In the figure, blue and red colors represent the parameters of worker $1$ ($x_1$) and $2$ ($x_2$), respectively. This illustration is provided for a model with $6$ layers ($6$ connected circles). The colors of the layers of the center model $x_c$ are chosen based on the closeness to the layers of the workers (or in other words the weights of the layers of the workers). In the case of MGRAWA (on the left), the parameters of the center model in different layers have the same color since all layers of the center model are influenced in the same way by each worker (each worker is assigned just a single weight). Furthermore, the layers of the central model are reddish, which implies that the weight coefficient for $x_2$ was greater than for $x_1$ and thus the $2^{\text{nd}}$ worker had more influence on the center model than the $1^\text{st}$ one. In LGRAWA (on the right), layers of the center model do not have the same color, unlike in MGRAWA. This is because in LGRAWA the weighted averaging is applied on the layer level, not the model level. Furthermore, the colors in the layers of $x_C$ are determined by their closeness to the corresponding layers of the workers. The layers that are closer to worker $1$ are more blueish-purplish whereas the ones closer to worker $2$ are more reddish-pinkish.

\subsection{Proximity Search Mechanism and Momentum Update}
The distributed update for both MGRAWA and LGRAWA requires more time than the other distributed training algorithms due to the gradient accumulation phase. To compensate for that we select higher communication periods in order to balance the time allocated for local and distributed optimization. As a result, the flatness-encouraging updates for MGRAWA and LGRAWA algorithms are applied less frequently, which is detrimental for performance. To overcome this limitation, inspired by a similar mechanism in \citep{lsgd_paper}, we apply an additional force in the local optimization phase that pulls the workers toward the previously calculated center model. We refer to this additional force as the proximity search mechanism as it ensures that the workers do not completely move apart and encourages the flatter region of the center variable in the loss surface in the local optimization phase. We denote the proximity search coefficient by $\mu$ and scale it with the multiplicative inverse of the communication period $\tau$. To be more specific, the longer the communication period is, the more the worker is pulled towards an outdated direction. Thus, the effective applied force should  be $\frac{\mu}{\tau}$ so that the larger the $\tau$ is, the less proximity correction is applied. \\

Finally, the momentum update applied on the accumulated gradient scores stabilizes the flatness-seeking mechanism and helps to avoid the situation when a worker traversing on an otherwise smooth path, incidentally scores a high gradient norm value for the current selection of the subset of train samples and is therefore granted a small weight in the weighted average of MGRAWA and LGRAWA.

\section{Experiments}
\label{sec:Experiments}

\begin{table*}[ht]
\caption{Mean and standard deviation of the test errors (lower is better).}
\renewcommand{\arraystretch}{1.2}
\centering
\resizebox{0.9\textwidth}{!}{
\begin{tabular}{c|c|cccccc}
Experiment                                                                            & Model      & EASGD        & LSGD         & MGRAWA       & LGRAWA                & DP+SGD       & DP+SAM       \\ \hline
\multirow{3}{*}{\begin{tabular}[c]{@{}c@{}}CIFAR-10\\ 4 Workers\end{tabular}}  & ResNet-20  & $9.22_{\pm 0.20}$  & $9.23_{\pm 0.16}$  & $8.99_{\pm 0.27}$  & $\mathbf{8.93_{\pm 0.20}}$  & $9.67_{\pm 0.06}$  &   $10.91_{\pm 0.05}$           \\
                                                                                      & VGG-16     &  $7.68_{\pm 0.25}$   & $7.65_{\pm 0.13}$  &  $7.64_{\pm 0.15}$   &  $\mathbf{7.52_{\pm 0.13}}$   &  $8.67_{\pm 0.06}$   & $8.92_{\pm 0.39}$             \\
                                                                                      & PyramidNet & $4.06_{\pm 0.16}$  & $3.98_{\pm 0.05}$  & $4.04_{\pm 0.11}$  & $\mathbf{3.79_{\pm 0.08}}$  & $4.19_{\pm 0.07}$  &  $5.54_{\pm 0.06}$     \\ \hline
\multirow{3}{*}{\begin{tabular}[c]{@{}c@{}}CIFAR-100\\ 8 Workers\end{tabular}}       & DenseNet  &  $23.61_{\pm 0.32}$  &  $\mathbf{22.55_{\pm 0.14}}$  &  $22.82_{\pm 0.28}$   &    $22.72_{\pm 0.22}$     &  $23.42_{\pm 0.07}$  &  $28.48_{\pm 0.09}$  \\ 
                                                                                    & WideResNet & $20.37_{\pm 0.15}$ & $20.01_{\pm 0.21}$ & $\mathbf{19.68_{\pm 0.13}}$  &  $19.71_{\pm 0.02}$   &  $21.47_{\pm 0.08}$  &  $29.54_{\pm 0.07}$  \\
                                                                                      & PyramidNet & $19.02_{\pm 0.06}$ & $19.25_{\pm 0.12}$ & $19.05_{\pm 0.14}$ & $\mathbf{18.82_{\pm 0.17}}$ & $19.59_{\pm 0.25}$ & $29.85_{\pm 0.96}$ \\ \hline

ImageNet(12W)  & ResNet-50 & $28.59_{\pm0.05}$        & $26.17_{\pm0.06}$        &  $\mathbf{25.35_{\pm0.04}}$    &    $ 25.68_{\pm0.08}$  &  $25.49_{\pm0.04}$   &   $36.92_{\pm0.11}$ \\                                                                                    
\end{tabular}}
\label{main_body:test_errors}
\end{table*}

\begin{table*}[ht]
\caption{Mean and standard deviation of the Frobenius norm of the Hessian matrices (lower is better).}
\renewcommand{\arraystretch}{1.2}
\resizebox{\textwidth}{!}{
\begin{tabular}{c|c|cccccc}
Experiment & Model        & EASGD          & LSGD          & MGRAWA                  & LGRAWA         & DP+SGD         & DP+SAM          \\ \hline
\multirow{3}{*}{\begin{tabular}[c]{@{}c@{}}CIFAR-10 \\ 4 Workers\end{tabular}} &  ResNet-20  &  $\mathit{84.12_{\pm 5.32}}$  & $317.96_{\pm 4.55}$  &   $84.32_{\pm 3.72}$   & $\mathbf{78.13_{\pm 2.06}}$ &  $324.97_{\pm 23.79}$ &  $113.45_{\pm 5.41}$ \\
& VGG-16     & $258.42_{\pm 40.50}$ &   $252.70_{\pm 17.94}$   &  $\mathit{199.18_{\pm 9.47 }}$   &  $227.03_{\pm 10.36 }$  &   $864.52_{\pm 73.77}$  &   $\mathbf{134.81_{\pm 2.38 }}$              \\ 
& PyramidNet & $341.28_{\pm 35.97}$      & $526.61_{\pm 73.22}$      & $\mathit{297.59_{\pm 48.31}}$       &   $\mathbf{240.06_{\pm 59.12}}$     &  $474.26_{\pm 53.80}$      & $318.55_{\pm 68.38}$      \\ \hline

\multirow{3}{*}{\begin{tabular}[c]{@{}c@{}}CIFAR-100 \\ 8 Workers\end{tabular}} & DenseNet   & $194.04_{\pm 12.98}$ & $194.65_{\pm 3.67}$ & $\mathbf{164.89_{\pm 25.65}}$ & $\mathit{177.29_{\pm 10.13}}$ & $435.74_{\pm 49.55}$ & $633.689_{\pm 47.30}$ \\
& WideResNet & $964.11_{\pm 83.16}$      &  $1037.38_{\pm 126.20}$     &  $\mathbf{584.13_{\pm 77.42}}$      &  $708.94_{\pm 109.86}$        &   $1382.94_{\pm 234.10}$       &    $\mathit{647.11_{\pm 56.59}}$  \\
& PyramidNet &   $241.70_{\pm 31.38}$   & $750.07_{\pm 52.73}$   &  $\mathit{233.98_{\pm 14.82}}$    &  $255.47_{\pm 47.52}$  & $284.66_{\pm 26.87}$ &  $\mathbf{145.57_{\pm 5.38}}$ 

\end{tabular}}
\label{main_body:frob_hessian}
\end{table*}

\subsection{Experiment Details}
We consider four parameter-sharing distributed training optimizers: EASGD, LSGD, MGRAWA, and LGRAWA, and one gradient-sharing method DistributedDataParallel (DP) \citep{data_parallel} coupled with the SGD optimizer. To show the performance of the well-known flat-minima-seeking optimizer SAM, we couple it with the DP method, the vanilla distributed training algorithm. We would like to emphasize that we conduct the performance analysis of the listed methods in a time-constrained training environment, that is, the training is limited by the total training time, not the number of epochs since main motivation behind using data-parallel distributed training is to reduce the overall training time. Our code base is publicly available on \href{https://github.com/tolgadimli/GRAWA}{\texttt{github.com/tolgadimli/GRAWA}}. \\

\vspace{-3mm}
\begin{figure}[ht]
    \centering
    \includegraphics[width=0.8\columnwidth]{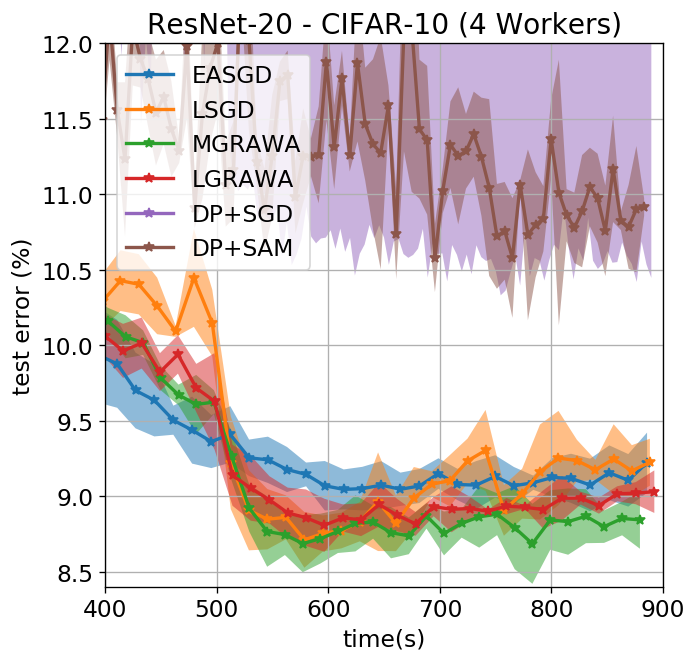}
    \caption{Test error (\%) curves of different distributed training methods in 4 workers and ResNet-20 setting.}
\end{figure}

We carry out experiments in three different settings with varying numbers of workers: $4$, $8$, and $12$. We use the well-known image classification data sets CIFAR-10 \citep{cifar10}, CIFAR-100 \citep{cifar100} and ImageNet \citep{imagenet}. In the experiments with CIFAR-10 and $4$ workers, we use ResNet-20 \citep{resnet}, VGG16 \citep{vgg}, PyramidNet-110 ($\alpha=270$) \citep{pyramidnet} models, and in the experiments with CIFAR-100 and $8$ workers, we use DenseNet-121 \citep{densenet}, WideResNet \citep{wideresnet} type WRN28-10 without the bottleneck layer, and PyramidNet again.  Finally, in the ImageNet experiments, we utilized all $12$ workers (12W) in our distributed training environment and trained ResNet-50 models. The other details regarding the training procedure can be found in Appendix \ref{appendix:experiment_details}. \\




\subsection{Experiment Results}

In this subsection, we present the final test errors and flatness proxy values that are obtained using our proposed LGRAWA and MGRAWA methods and the other competitor methods namely, LSGD, EASGD, and the vanilla algorithm \textit{DistributedDataParallel} which is also coupled with the SAM optimizer. As can be seen from Table \ref{main_body:test_errors}, GRAWA family algorithms achieve the smallest errors in most settings consistently. Particularly, the LGRAWA method keeps its superiority in the majority of the results. \\

\vspace{-5mm}
\begin{figure}[ht]
    \centering
    \includegraphics[width=0.8\columnwidth]{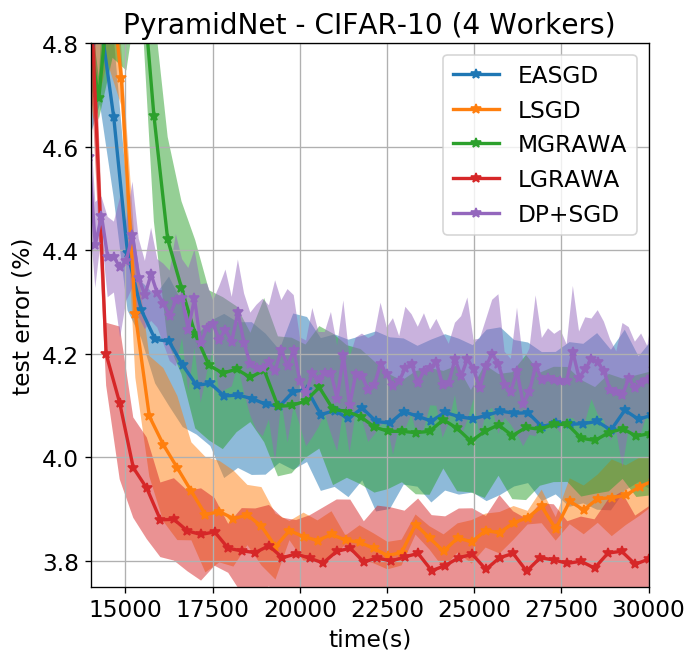}
    \caption{Test error (\%) curves of different distributed training methods in 4 workers and PyramidNet setting.}
\end{figure}

We also carry out flatness analysis with the eigenvalues of the Hessian matrix. To do this, we utilize the Lanczos algorithm \citep{lanczos1, lanczos2} in the first 100 most representative directions of the Hessian Spectra. Then, we approximate the Frobenius norm of the Hessian matrix, a good surrogate of the sharpness \citep{fantastic_measures}, by using the extracted 100 pseudo-eigenvalues. The results are provided in the table \ref{main_body:frob_hessian}. Note that we exclude the Imagenet experiments from this set of results due to the dataset size. The results in the table show that MGRAWA and LGRAWA algorithms encourage flat minima consistently while ensuring smaller test errors. Notice that in two settings the SAM optimizer yields a flatter minimum but its corresponding test errors are considerably higher than the GRAWA family. Finally in Table \ref{main_body:comm_times} from appendix, we show that the GRAWA family algorithms require less amount of inter-worker communication time than its competitors.

\begin{table*}[ht]
\centering
\caption{Scalability analysis results for MGRAWA and LGRAWA}
\vspace{2mm}
\renewcommand{\arraystretch}{1.2}
\resizebox{0.85\textwidth}{!}{\begin{tabular}{c|ccc|ccc}
\multicolumn{1}{l|}{}  & \multicolumn{3}{c|}{MGRAWA}             & \multicolumn{3}{c}{LGRAWA}                \\ \hline
Model - Dataset        & 4 Workers             & 8 Workers      & 12 Workers         & 4 Workers               & 8 Workers      & 12 Workers   \\ \hline
ResNet-20 - CIFAR-10   & $8.99_{\pm 0.27}$  & $8.79_{\pm 0.19}$ & $8.84_{\pm 0.13}$  & $8.93_{\pm 0.20}$    & $8.97_{\pm 0.16}$  & $8.94_{\pm 0.11}$\\
PyramidNet - CIFAR-100 & $19.05_{\pm 0.14}$ & $19.00_{\pm 0.08}$ & $19.07_{\pm 0.05}$  & $18.94_{\pm 0.07}$ & $18.82_{\pm 0.17}$ & $18.86_{\pm 0.11}$
\end{tabular}}
\label{table:scalability}
\end{table*}

\subsection{Ablation Study}

We also carry out an ablation study to observe how our algorithm performs in the same setting when we increase the number of workers participating in the distributed training environment. Particularly, we test the scalability of the MGRAWA and LGRAWA methods by training ResNet-20 models on CIFAR-10 dataset and PyramidNet models on CIFAR-100 with 4, 8 and 12 workers. For each model-dataset-worker combination, the experiments are run for 3 different seeds, the resulting mean and standard deviation values are reported in Table \ref{table:scalability}. \\

Our scalability analysis reveals that the proposed algorithms MGRAWA and LGRAWA do not suffer from performance degradation when the number of workers in the training environment is increased. We also argue that it might be possible to benefit from an increased number of workers in the distributed setting as more information about the loss landscape can be collected by more workers traversing and exploring it. This is left as an open question and it requires further investigation. \\

\vspace{-5mm}
\begin{figure}[ht]
    \centering
    \includegraphics[width=0.8\columnwidth]{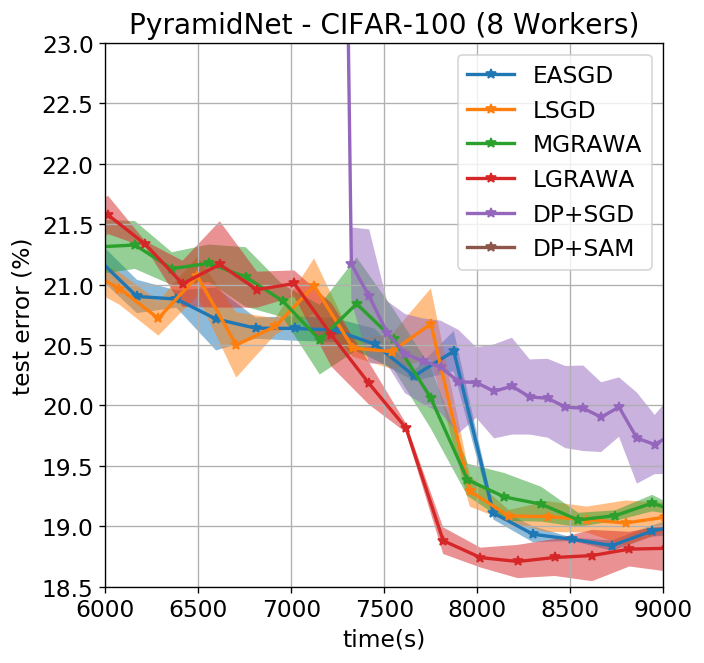}
    \caption{Test error (\%) curves of different distributed training methods in 8 workers and PyramidNet setting.}
\end{figure}

\section{Conclusion}
\label{sec:Conclusion}
\vspace{-2mm}
In this paper, we propose a novel asynchronous algorithm family, which are namely MGRAWA and LGRAWA, for the distributed training of deep learning models. They fall into the category of knowledge transfer methods with periodic parameter sharing, similar to the LSGD and EASGD. In the distributed update phase, our algorithms employ a weighted averaging scheme. In this scheme, the weights are determined inversely proportional to the accumulated gradients inside the network layers which signals the smoothness of the model's trajectory on the loss valley and, how mature the model components are. Our motivation behind designing such a weighted averaging scheme is grounded on the importance of the flatness of the loss valley for the generalization capability of the final model, as well as prioritizing the network layers that require less correction, namely mature components. We provide theoretical proof for the convergence rate of the vanilla GRAWA algorithm in the convex setting and provide a convergence analysis in the non-convex case that holds for GRAWA and both of its variants. We also demonstrate the effectiveness of our algorithms MGRAWA and LGRAWA empirically by training deep learning models with various architectures and on different data sets. Our experimental results show that MGRAWA and LGRAWA accelerate the convergence rate and achieve lower error rates than SOTA competitors while yielding flatter minima and requiring less communication. The scalability of the proposed algorithms is also investigated and no performance drop is observed when the number of workers is increased. \\

\vspace{-3mm}
\section{Acknowledgement}
The authors acknowledge that the NSF Award \#2041872 sponsored the research in this paper.

\bibliography{references}
\bibliographystyle{abbrvnat}

\appendix
\onecolumn

\section{Motivational Example}
\label{appendix:motivational_example}
As mentioned in section the motivational example section, we use the Vincent function to demonstrate the flatness encouraging updates of the GRAWA algorithm and its variants MGRAWA and LGRAWA. Loss surface of this function ($f(x,y) = - \sin \left ( 10\ln (x) \right ) - \sin \left ( 10\ln (y) \right )$) is provided in the figure below.

\begin{figure}[H]
\centering
\includegraphics[width=0.45\columnwidth, trim={2.5cm 2cm 0cm 2.5cm}, clip]{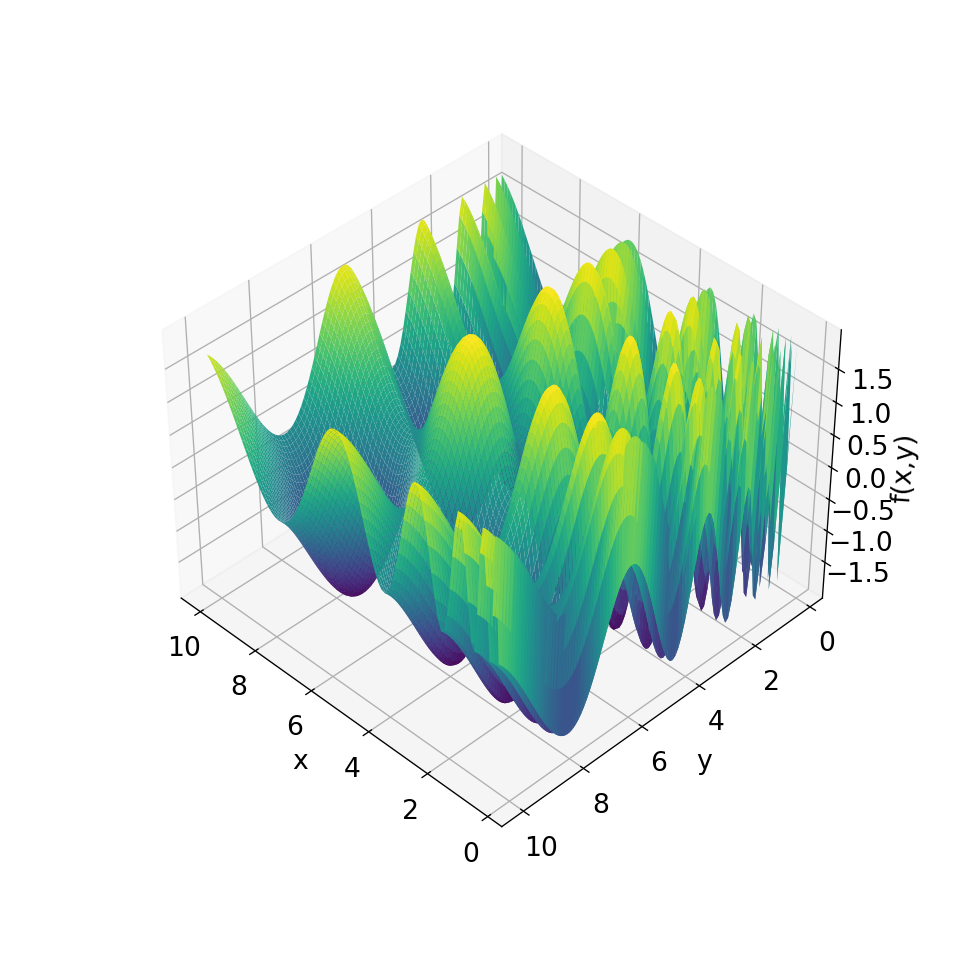}
\caption{Loss surface of the Vincent function with 2-dimensional input}
\end{figure}

We initialized $4$ workers from the corners of the $(x,y)$ plane above. More specifically, workers are initialized from the following points: ${(0.25, 0.25), (0.25, 10), (10, 0.25), (10,10)}$. Each worker runs the vanilla SGD algorithm as its local optimizer with a learning rate of $0.01$ and then it is coupled with a distributed optimizer that applies a pulling force after every 4 local updates in accordance with the policy of the distributed training algorithm. We used 5 different distributed optimizers which are EASGD, LSGD, GRAWA, MGRAWA, and LGRAWA. In the following figures, we share the trajectories followed by all of the workers when each of the aforementioned distributed optimizers is used. \\

\begin{figure}[H]
\centering
\includegraphics[width=0.85\columnwidth, trim={2cm 0cm 2cm 0.2cm}, clip]{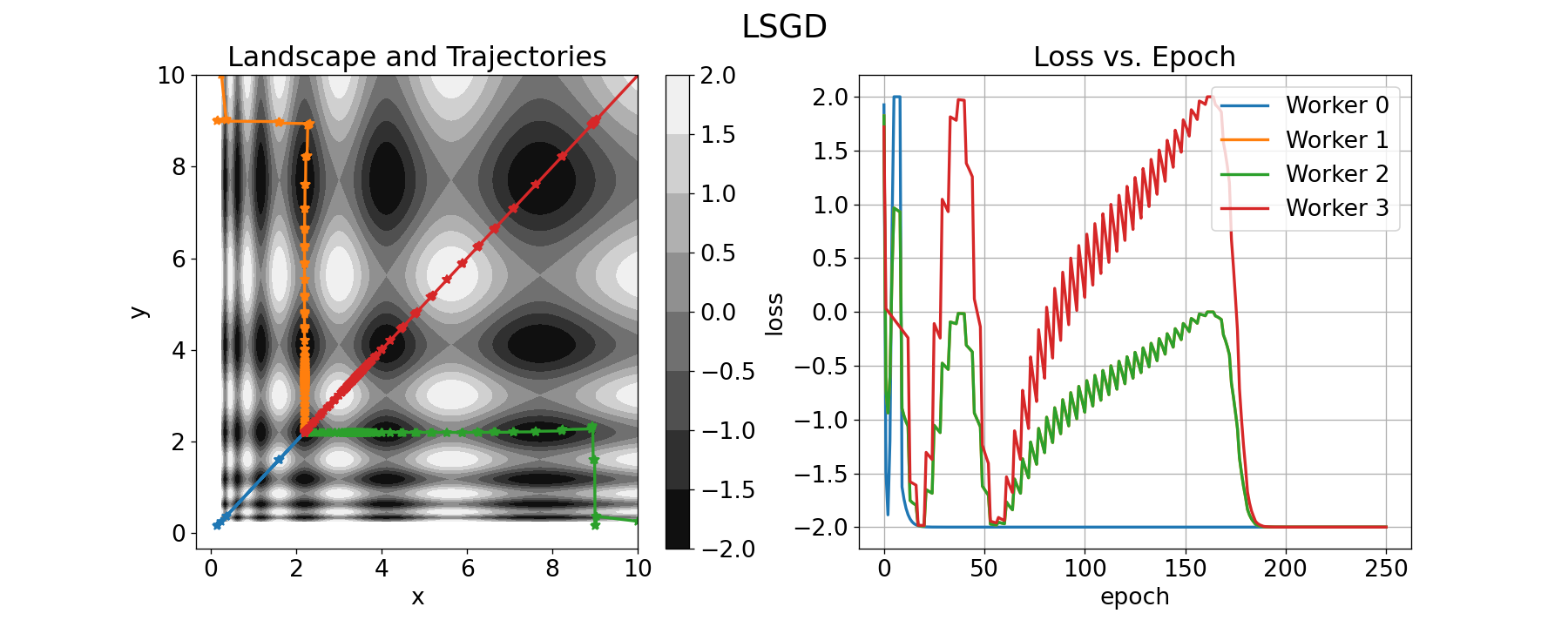}
\caption{Trajectories of the workers and when the distributed optimizer is LSGD}
\end{figure}

\begin{figure}[H]
\centering
\includegraphics[width=0.85\columnwidth, trim={2cm 1cm 2cm 0.2cm}, clip]{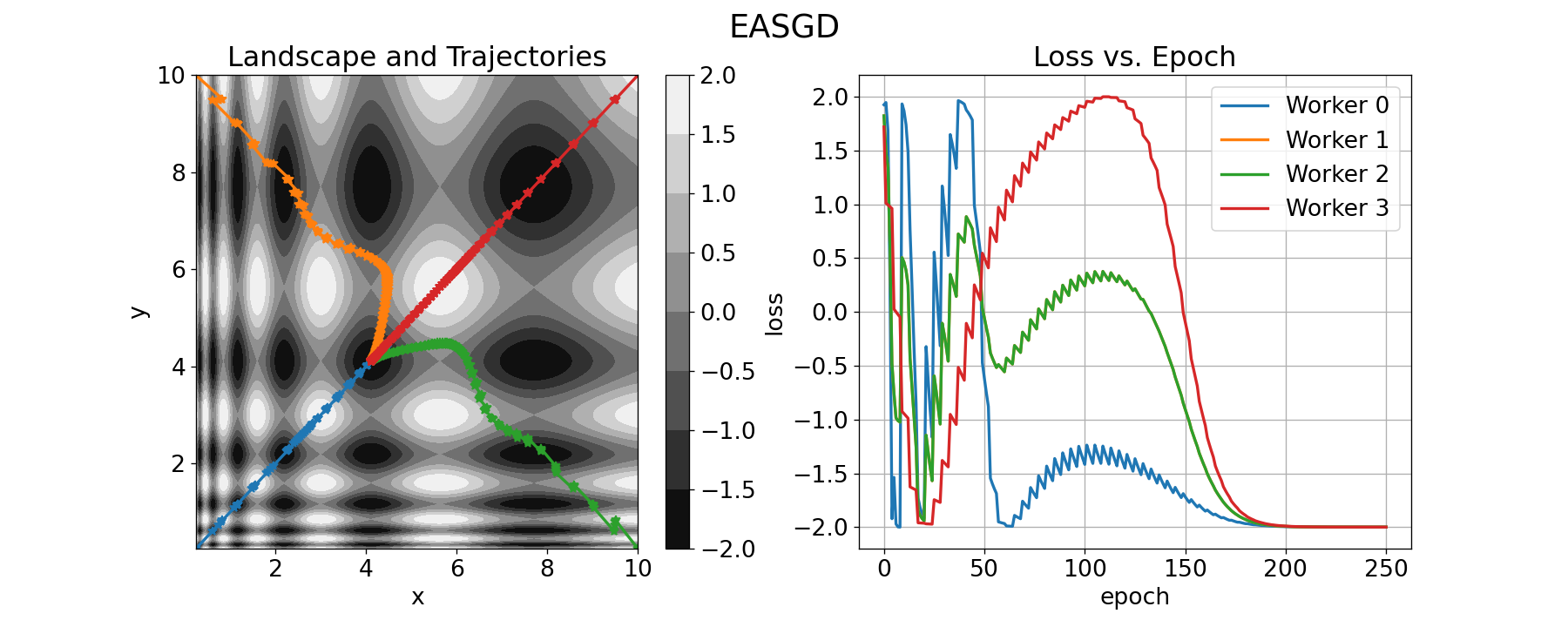}
\caption{Trajectories of the workers and when the distributed optimizer is EASGD}
\end{figure}

\begin{figure}[H]
\centering
\includegraphics[width=0.85\columnwidth, trim={2cm 1cm 2cm 0.2cm}, clip]{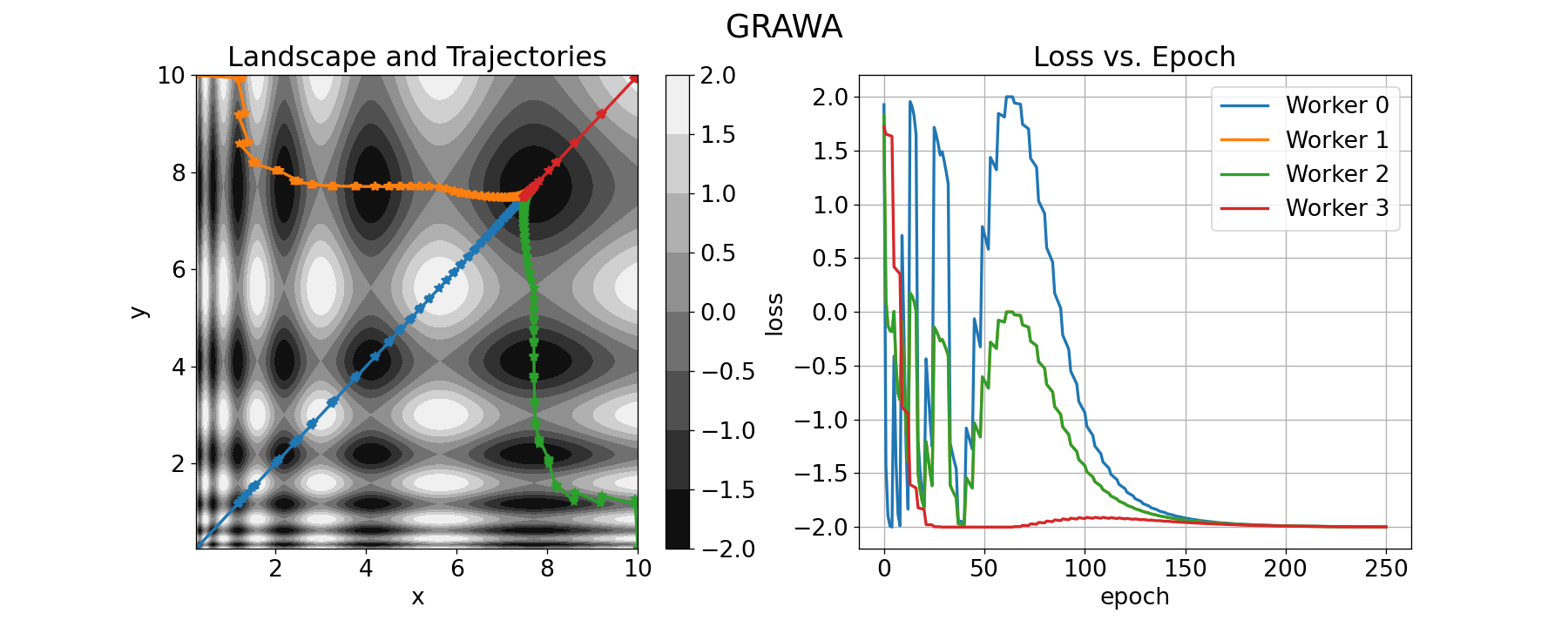}
\caption{Trajectories of the workers and when the distributed optimizer is GRAWA}
\end{figure}

\begin{figure}[H]
\centering
\includegraphics[width=0.85\columnwidth, trim={2cm 1cm 2cm 0.2cm}, clip]{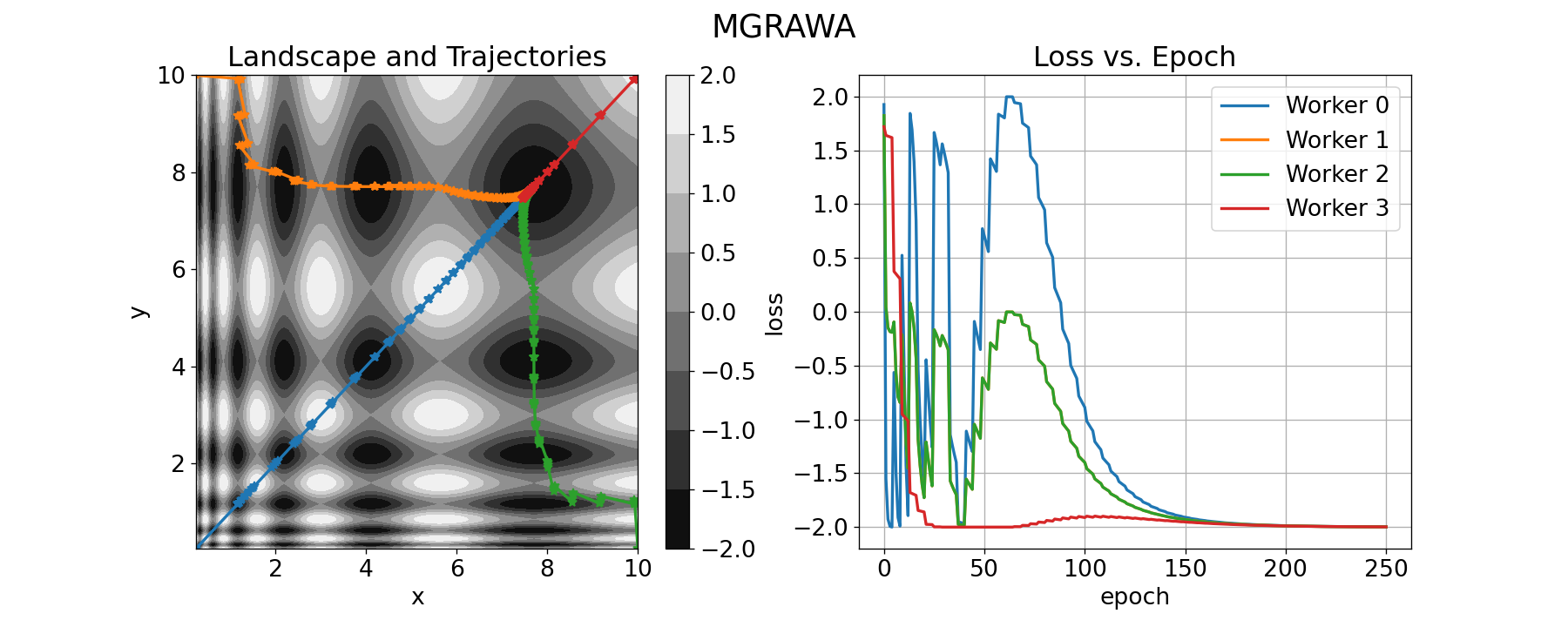}
\caption{Trajectories of the workers and when the distributed optimizer is MGRAWA}
\end{figure}

\begin{figure}[H]
\centering
\includegraphics[width=0.85\columnwidth, trim={2cm 1cm 2cm 0.2cm}, clip]{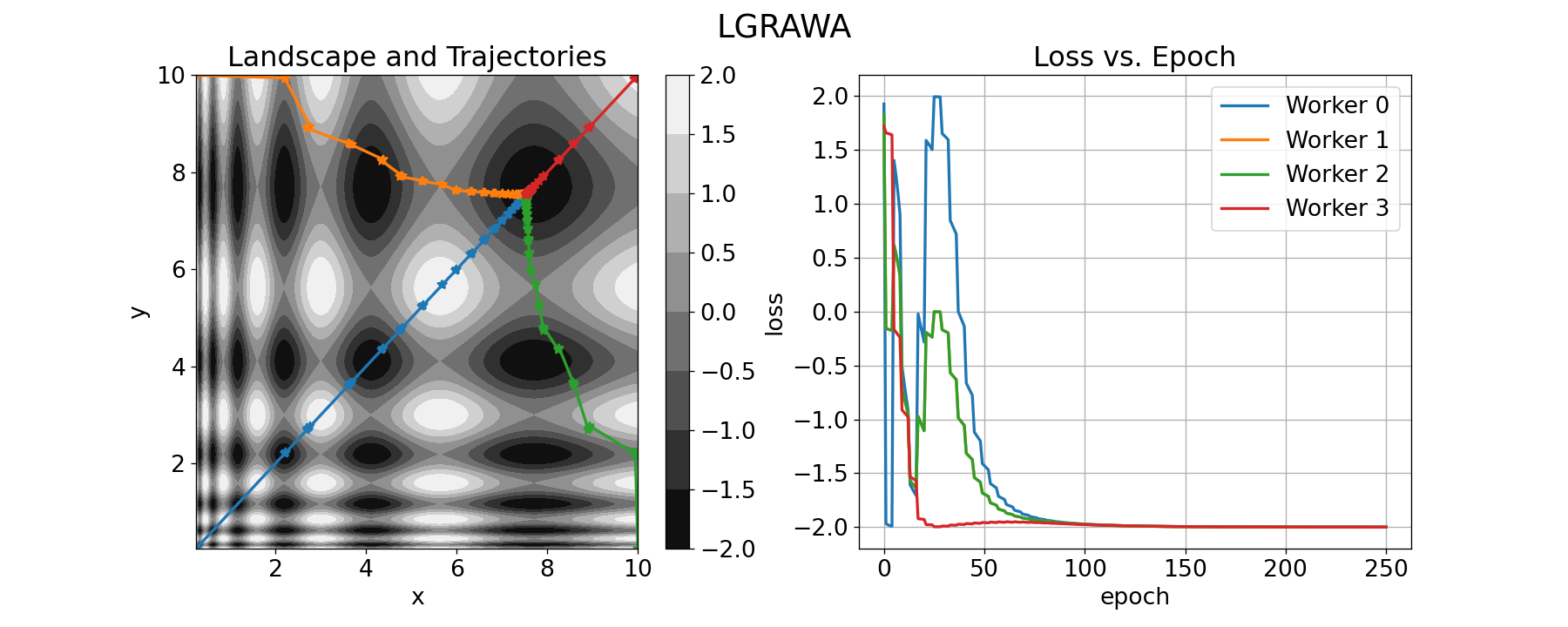}
\caption{Trajectories of the workers and when the distributed optimizer is LGRAWA}
\end{figure}

As can be seen from the trajectory figures above, although LSGD manages to spot the minimizer early but with a sharp valley, which results in a low generalization capability. It constantly pulls all of the workers into that valley because the leader is trapped there. As for the EASGD algorithm, independent of the geometry of the loss surface and performance of the other workers, it constantly pulls all the workers towards the average of them and finally converges to a valley with a slightly wider valley than LSGD. In the case GRAWA and its variants (MGRAWA and LGRAWA), the distributed optimization process constantly pushed the workers toward the flatter region of the valley, converging to a point with a small loss value and good generalization capability.

\section{Experiment Details}
\label{appendix:experiment_details}

\subsection{Gradient Norm Experiments}
\label{appendix:grad_norm_exps}
In order to validate our hypothesis about the correlation between the gradient norm calculated on the final model parameters and the generalization gap (test error$(\%)$ - train error$(\%)$), we first train $162$ different ResNet-18 models with varying hyperparameters and architectures. The hyperparameters that are varied and their ranges can be found in the table below.

\vspace{-2mm}

\renewcommand{\arraystretch}{1.3}
\begin{table}[H]
\centering
\caption{Hyperparameter search space used in the Gradient Norm experiments}

\begin{tabular}{c|c}
\textbf{Hyperparameter}   & \textbf{Search Grid}      \\ \hline
Number of Residual Layers & $[4,6,8]$             \\
Momentum Coefficient      & $[0.1, 0.5, 0.9]$      \\
Learning Rate             & $[0.01, 0.0075, 0.005]$ \\
Batch Size                & $[32, 128, 512]$        \\
Weight Regularization     & $[0, 0.0001]$     
\end{tabular}
\end{table}

\vspace{-5mm}

In total there are 162 different models trained with different hyperparameters. For training the models, no augmentation technique is applied on the train set. Only normalization is applied as pre-processing with the mean vector $(0.4914, 0.4822, 0.4465)$ and the standard deviation vector $(0.2023, 0.1994, 0.2010)$. Also note that both train errors and the gradient norms are calculated on the whole train set of the CIFAR-10 data without any augmentation. Similarly, test errors are calculated on the whole test set.

\subsection{Distributed Training Experiments}

\subsubsection{Data Preprocessing and Loading}

Since we are using the PyTorch-based distributed training library introduced with the LSGD paper, we stick to the same data preprocessing steps. In all the experiments with the CIFAR-10 and CIFAR-100 dataset, images are normalized with the mean vectors $(0.4914, 0.4822, 0.4465), (0.5071, 0.4867, 0.4408)$ and the standard deviation vectors $(0.2023, 0.1994, 0.2010), (0.2675, 0.2565, 0.2761)$ respectively. To augment the training set, a horizontal flip with a probability of $0.5$ is used. For the ResNet-20 model, the original $3 \times 32 \times 32$ CIFAR-10 images are randomly cropped to be presented as images of size $3 \times 28 \times 28$ to the network. Note that, for the test samples, the $3 \times 28 \times 28$ part in the center of the image is extracted. 

For the VGG-16, PyramidNet, DenseNet, and WideResNet models, train set samples are first padded to the size $3 \times 40 \times 40$, then $3 \times 32 \times 32$ portion of the padded image is randomly cropped and presented to the network for augmentation. The testing is carried out with the original $3 \times 32 \times 32$ images without any preprocessing except the normalization.

In the ImageNet experiment with ResNet-50 model, we normalize samples with the mean vector $(0.485, 0.456, 0.406)$ and the standard deviation vector $(0.229, 0.224, 0.225)$. For the augmentation of train set, the image portions are randomly crop such that the crop accounts an area between $8\%$ or $100\%$ of the original image size and then it is resized to the shape $3 \times 224 \times 224$. We also randomly apply a horizontal flip. For the test samples, the images are resized to $3 \times 256 \times 256$, and the center $3 \times 224 \times 224$ part is cropped.

To load the data, PyTorch's distributed data sampler is used. It automatically splits the data into non-overlapping data shards and each worker is only trained with its own shard. Notice that this only applies to the train portion of the dataset and the number of data shards is equal to the number of workers in the parallel computing environment.

\subsubsection{Local Optimizer Details}
 Unless otherwise stated, in all of the experiments we use SGD with a learning rate of $0.1$, a momentum of $0.9$ and Nesterov acceleration as the local optimizer \cite{momentum}. The weight decay values are selected and learning rate drops are applied in accordance with the training details in the aforementioned papers describing the vision models we use. For the SAM optimizer, we select the base optimizer as SGD and we use the suggested value in the paper for the neighborhood search parameter, that is $\rho = 0.05$.

\subsubsection{Hyperparameter Search Grids}
\label{appendix:HyperParam}

In the tables below, the hyperparameter search grid for different distributed training optimizers is provided for the experiments on the CIFAR-10 and CIFAR-100 datasets.

\begin{table}[H]
\caption{Hyperparameter search space for LSGD, EASGD, MGRAWA, and LGRAWA}
\centering
\renewcommand{\arraystretch}{1.3}
\begin{tabular}{c|c|c|c}
              & LSGD                & EASGD                         & MGRAWA \& LGRAWA        \\ \hline
Comm. Period$(\tau)$  & 4, 8, 16        & 4, 8, 16                 & 16, 32        \\
Pulling Force$(\lambda)$ & 0.01, 0.1 & 0.1, 0.3, 0.43*              & 0.3, 0.5, 0.7\\
Prox. Search$(\mu)$  & 0.01, 0.1        & NA                            & 0.01, 0.05                 
\end{tabular}
\end{table} 

For each of the optimizers mentioned in the table above, the learning rate of the local SGD optimizer is set to $0.1$ and its momentum value is 0.9. Also, a fixed batch size of $128$ is used. Note that, for EASGD, pulling force of $0.43$ is specifically added to the hyperparameter search space since it is mentioned in the original EASGD paper.

\begin{table}[H]
\caption{Hyperparameter search space for DataParallel}
\centering
\renewcommand{\arraystretch}{1.3}
\begin{tabular}{c|c}
              & DataParallel          \\ \hline
Learning Rate & 0.01, 0.1 \\
Batch Size    & 32, 64, 128          
\end{tabular}
\end{table} 

Because the DataParallel algorithm processes the gradient from each worker and applies the same update to the all workers, it does not transfer the knowledge with parameter sharing. Hence, we do not have a communication period or a pulling force towards the center model. Instead, we vary the local learning rate of the optimizer and the batch size.

For the ImageNet experiments using the ResNet-50 model, we have used the same hyperparameter search grids for the parameter sharing methods as before. In the 12 worker setting, LSGD has global and local communication periods. Following the convention in the paper LSGD paper, the local communication period is picked as a quarter of the global communication period. $\tau_L = \frac{\tau_G}{4}$. For the DataParallel algorithm, the fixed batch size of $32$ is used and only the learning rate is varied among the values ${0.01, 0.1}$.

\section{Hardware Details}
We run both the flatness experiments and carry out the distributed training of deep learning models on 3 machines, each of which has 4 $\times$ NVIDIA GTX 1080 GPU cards (in total 12 GPUs). The machines are connected over Ethernet.

\section{Additional Experiment Results}

In the table below, we report the total time spent on the communication for each parameter-sharing distributed optimizer, in their hyperparameter setting that yields the lowest test error. As can be seen, the GRAWA family requires the least amount of time spent on inter-worker communication.

\begin{table*}[ht]
\caption{Time spent on inter-worker communication in seconds(s) and total number of inter-worker communications (in parenthesis) for the parameter sharing distributed training methods (lower is better).}
\renewcommand{\arraystretch}{1.1}
\resizebox{0.9\textwidth}{!}{
\begin{tabular}{c|c|cccc}
Experiment                                                                            & Model      & EASGD        & LSGD         & MGRAWA     & LGRAWA     \\ \hline
\multirow{3}{*}{\begin{tabular}[c]{@{}c@{}}CIFAR-10\\ 4 Workers\end{tabular}}  & ResNet-20  & 57s (2539)   & 29s (1319)   & \textbf{26s (167)}  & 50s (328)  \\
                                                                                      & VGG-16     & 100s (2747)  & 95s (2783)   & 81s (353)  & \textbf{75s (351)}  \\
                                                                                      & PyramidNet & 3673s (15053)  & 3252s (14842)     & \textbf{1602s (1858)}     &  1608s (1886) \\ \hline
\multirow{3}{*}{\begin{tabular}[c]{@{}c@{}}CIFAR-100\\ 8 Workers\end{tabular}} & DenseNet   & 590s (7557)  & 438s (6761)  & \textbf{291s (927)} &   317s (923)        \\
                                                                                      & WideResNet & 880s (2991)  & 2268s (2469) & 531s (392) & \textbf{435s (393)} \\
                                                                                      & PyramidNet & 1884s (7381) & 1177s (7496) & \textbf{730s (979)} & 767s (971)
\end{tabular}}
\centering
\label{main_body:comm_times}
\end{table*}

We also provide additional plots that include convergence curves of the other four settings in Table \ref{main_body:test_errors} that are not presented in the main body due to page limit.



\begin{figure}[h!]
\centering
\subfigure{\includegraphics[width=0.45\textwidth]{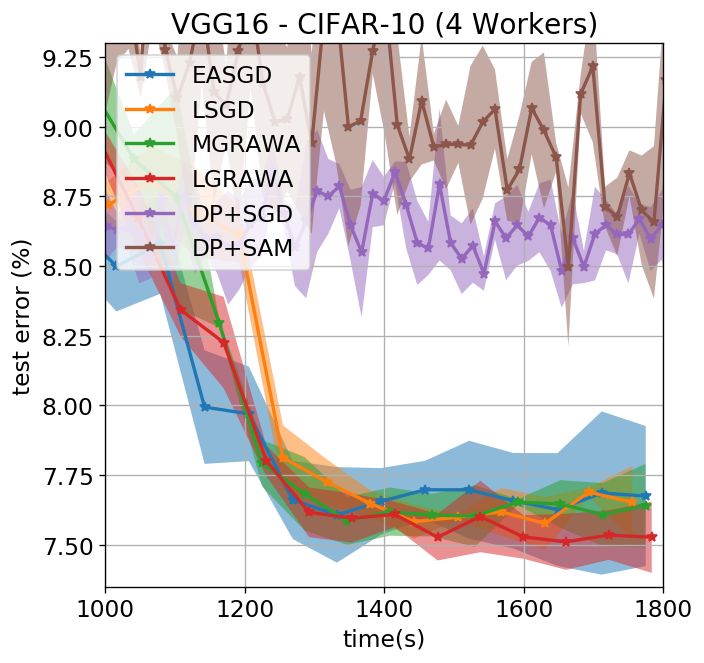}}
\hfill
\subfigure{\includegraphics[width=0.45\textwidth]{appendix_figs/pyramidnet-cifar10-4w.png}}
\hfill
\end{figure}

\begin{figure}[h!]
\centering
\subfigure{\includegraphics[width=0.45\textwidth]{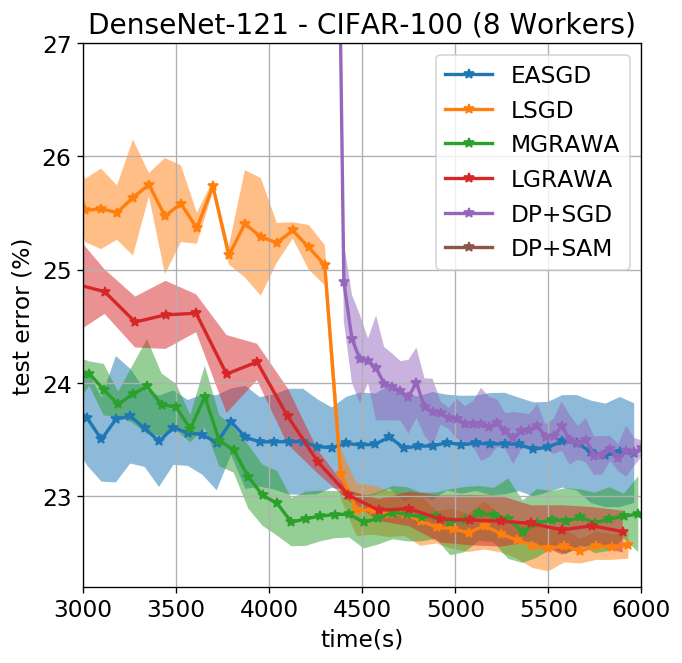}}
\hfill
\subfigure{\includegraphics[width=0.45\textwidth]{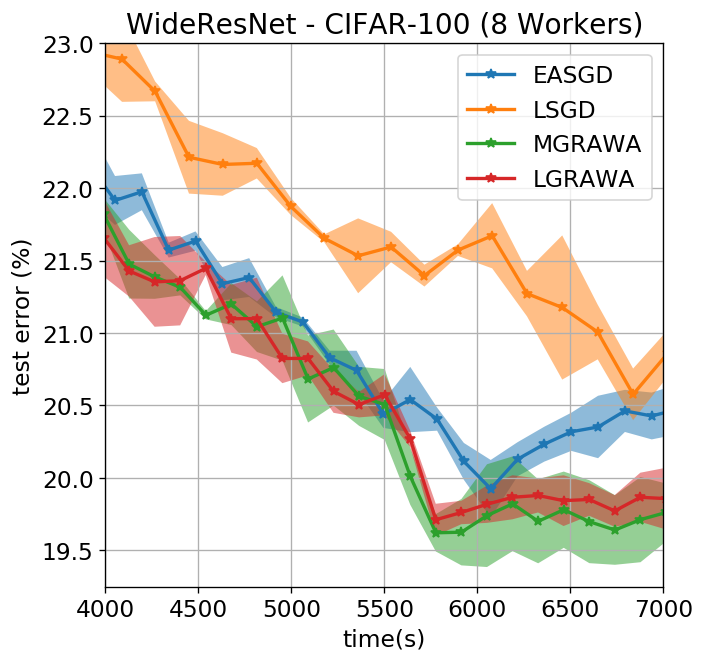}}
\hfill
\end{figure}

We also report the generalization gap that is simply calculated as test error (\%) - train error (\%). We only show cases when the generalization gap is not equal or very close to the test error and we observe that LGRAWA attains the smallest gap overall. 

\begin{table}[h!]
\centering
\resizebox{1\linewidth}{!}{
\begin{tabular}{c|c|cccccc}
Experiment                                                                       & Model      & EASGD & LSGD & MGRAWA & LGRAWA & DP+SGD & DP+SAM \\ \hline
\multirow{2}{*}{\begin{tabular}[c]{@{}c@{}}CIFAR-10\\4 Workers\end{tabular}} & ResNet-20  & $6.75_{\pm 0.32}$  & $6.83_{\pm 0.27}$  & $6.50_{\pm 0.24}$  & $\mathbf{6.39_{\pm 0.16}}$  & $7.45_{\pm 0.59}$  & $9.35_{\pm 0.52}$      \\
                                                                                 & VGG-16    & $7.08_{\pm 0.26}$  & $7.19_{\pm 0.17}$  & $6.86_{\pm 0.15}$  & $\mathbf{6.22_{\pm 0.34}}$   & $8.38_{\pm 0.30}$  & $8.59_{\pm 0.54}$     \\ \hline
CIFAR-100 (8W)                                                                   & DenseNet & $23.53_{\pm 0.31}$ & $22.49_{\pm 0.18}$ & $22.49_{\pm 0.45}$ & $\mathbf{22.11_{\pm 0.59}}$ & $23.42_{\pm 0.07}$ & $28.35_{\pm 0.26}$      
\end{tabular}}
\end{table}
\section{Details of the Theoretical Results}

\subsection{Technical Preliminaries }
\label{appendix:tech_prelim}

Recall the problem setting in a distributed training environment. There are $M$ workers $(x_1, x_2, ..., x_M)$, and there is a function $f$ (here we use $f$ instead of $F$) which is to be collectively minimized by all of the workers. In the GRAWA algorithm, there is a center variable $x_C$ that is obtained by a weighted average of all of the workers. Let $\beta_1, \beta_2, ..., \beta_M$ be the weights associated with the workers $x_1, x_2, ..., x_M$ respectively in this weighted averaging scheme. Also, recall that these weights are inversely proportional to the gradient norms of the function at points $x_1, x_2, ..., x_M$ and, $\sum_{i=1}^M \beta_i = 1$. This corresponds to the following expression for each weight $\beta_i$:

\begin{equation}
    \label{prelim:grad_weights}
    \beta_i = \frac{\Theta}{||\nabla f(x_i)||} \quad \text{where} \quad \Theta = \frac{ \prod_{i=1}^M ||\nabla f(x_i)||}{  \sum_{i=1}^M \frac{\prod_{j=1}^M ||\nabla f(x_j)||}{||\nabla f(x_i)||} }
\end{equation}

\subsection{Converge Analysis in Convex Case}

\textbf{Assumption:} Let us have a function $f$ with minimizer $x^*$ such that the following conditions are satisfied $\forall x,y \in$ \textbf{dom} $f$:

\begin{itemize}
    \item $L$-smooth: $||\nabla f(x) - \nabla f(y)|| \leq L ||x-y||$
    \item $m$-strongly convex: $ f(y) \geq f(x) + \nabla f(x)^T(y-x) + \frac{m}{2} ||y-x||^2 $
    \item $k$-cone: $|f(x) - f(x^*)| \geq k||x - x^*|| $ i.e. lower bounded by a cone which has a tip at $x^*$ and a slope $k$
    \item $\mu$-sPL: $||\nabla f(x)|| \geq \mu \left( f(x) - f(x^*)\right)$
\end{itemize}

We would like to highlight that in the last precondition, we use a modified version of the well-known Polyak inequality, with a stricter upper bound (without the square) which we refer to as \textit{sPL}. 

\begin{theorem}
\label{theorem:aleqb}
For a $\mu$-sPL, $L$-smooth and $k$-cone function $f$ with minimizer $x^*$, if $f(a) \leq f(b)$ and $k\mu \geq L$ then $ || \nabla f(a) || \leq || \nabla f(b) ||$ $\forall a,b \in \text{\normalfont dom} f$. In other words, the function $G(y) = || \nabla y ||$ is a monotonously increasing function when $y$ is a $\mu$-sPL, $k$-cone and $L$-smooth function and when $k\mu \geq L$.
    
\begin{proof}
From the definition of $\mu$-sPL, we can write: $||\nabla f(b)|| \geq \mu \left( f(b) - f(x^*)\right) $. Because $f(a) \leq f(b)$, we can also write: $f(a) - f(x^*) \leq f(b) - f(x^*)$. Overall, this implies the following:

\begin{equation*} 
    \frac{1}{\mu} || \nabla f(b)|| \geq  f(a) - f(x^*)
\end{equation*}

Now, using the cone assumption, we can write $f(a) - f(x^*) \geq k ||a-x^*||$ which implies:

\begin{equation*}
    \frac{1}{\mu} || \nabla f(b)|| \geq  k ||a-x^*||
\end{equation*}

Then, from L-Lipschitz differentiable property, we have $||a-x^*|| \geq L^{-1}||\nabla f(a) ||$ which leads us to:

\begin{equation*}
     \frac{1}{\mu} ||\nabla f(b)|| \geq \frac{k}{L} || \nabla f(a) ||
\end{equation*}

So, for $k\mu \geq L$, we have:

\begin{equation*}
    || \nabla f(b)|| \geq \frac{k\mu}{L}|| \nabla f(a) || \geq || \nabla f(a) || \quad \text{which implies} \quad ||\nabla f(a)|| \leq ||\nabla f(b)||.
\end{equation*}

\end{proof}

\end{theorem}

\begin{theorem}
\label{theorem:grad_center_leq}
Let $x_1, x_2, ..., x_M \in \text{\normalfont dom} f$ and, $f$ is a convex, $\mu$-sPL, real-valued, continuous and $L$-Lipschitz differentiable function with. Let $x_C = \sum_{i=1}^{M} \beta_i x_i$ where each $\beta_i$ is calculated as in equation \ref{prelim:grad_weights} so that $\sum_{i=1}^{M} \beta_i = 1$. Then, $||\nabla f(x_C)|| \leq \sqrt{M} || \nabla f(x_i)||$ holds $\forall i \in {1, 2, ..., M}$.

\begin{proof}
Since $f$ is a continuous, real, convex function and $\sum_{i=1}^{M} \beta_i = 1$, we can write the Jensen's inequality as follows:

\begin{equation*}
    f \left (\sum_{i=1}^M \beta_i x_i \right)  \leq \sum_{i=1}^M \beta_i f(x_i) \quad \text{that is} \quad f(x_C) \leq \sum_{i=1}^M \beta_i f(x_i)
\end{equation*}

Then, by applying theorem \ref{theorem:aleqb}, we can write $|| \nabla f(x_C) ||^2 \leq  ||  \sum_{i=1}^M \beta_i \nabla f(x_i) ||^2$. Using the triangle-inequality of the norm, we can also write the following upper bound for $|| \nabla f(x_C) ||^2$.

\begin{equation*}
    || \nabla f(x_C) ||^2 \leq   \sum_{i=1}^M || \beta_i \nabla f(x_i) ||^2 = \sum_{i=1}^M \beta_i^2 || \nabla f(x_i) ||^2 \\
\end{equation*}

Plugging in the definition of $\beta_i$ in equation \ref{prelim:grad_weights}, we obtain:

\begin{equation*}
    || \nabla f(x_C) ||^2 \leq   \sum_{i=1}^M \frac{\Theta^2}{|| \nabla f(x_i) ||^2 } || \nabla f(x_i) ||^2 =  M \Theta ^ 2
\end{equation*}

Taking square root of both sides, we obtain:

\begin{equation}
    \label{grad_center_leq: final}
    || \nabla f(x_C) || \leq \sqrt{M} \Theta
\end{equation}

Because, each weight $\beta_i \in [0,1]$, we have:

\begin{equation*}
    \beta_i = \frac{\Theta}{|| \nabla f(x_i) ||} \leq 1 \quad \longrightarrow \quad \Theta \leq || \nabla f(x_i || \quad \forall i \in {1, 2,..., M}
\end{equation*}


By combining this with inequality \ref{grad_center_leq: final}, we can write:

\begin{equation*}
    || \nabla f(x_C) || \leq \sqrt{M} || \nabla f(x_i) || \quad \forall i \in {1, 2,..., M}
\end{equation*}

\end{proof}

\end{theorem}

\begin{theorem}
\label{appendix:final_leq}
Let $x_C = \sum_{i=1}^{M} \beta_i x_i$ and $\beta_i$'s are calculated as in equation \ref{prelim:grad_weights}. For an $L$-Lipschitz differentiable, $\mu$-sPL, $k$-cone real-valued continuous function $f$ with minimizer $x^*$ satisfies $f(x_C) \leq f(x_i)$ for all $i \in 1,2,...,M$ provided that $k\mu \geq L\sqrt{M}$. 

\begin{proof}

using the $k$-cone property, we can write:
\begin{equation*}
    f(x_i) - f(x^*) \geq k|| x_i - x^* || \quad \forall i \in 1, 2,..., M
\end{equation*}

From L-Lipschitz property, we know that  $L||x_i-x^*|| \geq  || \nabla f(x_i) ||$ and by combining with the above inequality we can write:
\begin{equation*}
    f(x_i) - f(x^*) \geq \frac{k}{L} || \nabla f(x_i) || \quad \forall i \in 1, 2,..., M
\end{equation*}

By making use of \ref{theorem:grad_center_leq}, we can write: 
\begin{equation*}
    f(x_i) - f(x^*) \geq \frac{k}{L\sqrt{M}} || \nabla f(x_C) || \quad \forall i \in 1, 2,..., M
\end{equation*}

Finally, by utilizing the $\mu$-sPL property, we reach the following inequality.
\begin{equation*}
    f(x_i) - f(x^*) \geq \frac{k}{L} || \nabla f(x_C) || \geq \frac{k\mu}{L\sqrt{M}} \left( f(x_C) - f(x^*)\right) \quad \forall i \in 1, 2,..., M
\end{equation*}

When $k\mu \geq L\sqrt{M}$, we can write: $f(x_i) \geq f(x_C)$.

\end{proof}

\end{theorem}

\begin{theorem}
Let $f$ be a function that holds \ref{appendix:final_leq}. Let $\tilde{g}(x)$ be an unbiased estimator for $\nabla f(x)$ with $\mathrm{Var}(\tilde{g}(x)) \leq \sigma^2 + \nu ||\nabla f(x)||^2$, and let $x_C$ be the center variable obtained with the GRAWA algorithm. Suppose that $\eta, \lambda$ satisfy $\eta \leq \left ( 2L(\nu + 1)  \right)^{-1}$ and $\eta \lambda \leq \frac{m}{2L}$, $\eta \sqrt{\lambda} \leq \frac{\sqrt{m} }{\sqrt{2}L}$. Then the GRAWA step satisfies:

\begin{equation*}
    \mathbb{E}[f(x^{t+1}) - f(x^*)] \leq (1-m\eta)(f(x^t) - f(x^*)) - \eta \lambda (f(x^t)-f(x_C)) + \frac{\eta^2 L}{2} \sigma^2
\end{equation*}

The presence of the new term due to the GRAWA update increases the speed of the convergence since $f(x_C) \leq f(x)$ from \ref{appendix:final_leq}. Then, $\limsup_{t \rightarrow \infty} \mathbb{E}[f(x^{t+1}) - f(x^*)] \leq \eta \frac{L}{2m} \sigma^2 $. If $\eta$ decreases at rate $\eta = O(\frac{1}{t})$, then $\mathbb{E}[f(x^{t+1}) - f(x^*)] \leq O(\frac{1}{t})$.

\begin{proof}
    The proof can be found in section 7.4 of the appendix of the LSGD paper.
\end{proof}

\end{theorem}

\subsection{Convergence Analysis in Non-Convex Scenario}

\begin{theorem}
\label{appendix:nonconvex}
Recall the main objective of the distributed training objective we have:

\begin{equation*}
\min_{x} \quad F(x; \mathcal{D}) = \min_{x_1,\dots, x_M} \sum_{m=1}^{M} \mathbb{E}_{\xi \sim \mathcal{D}_m } f (x_m; \xi) + \frac{\lambda}{2} || x_m - x_c ||^2
\end{equation*}

In our non-convex analysis, we make the following assumptions that impose $L$-smoothness condition on function F, bounds the variance of the calculated gradients using $\xi$ sampled from $\mathcal{D}_m$ for $m = 1, 2..., M$ and the norm of the Euclidean distance between the center variable and workers at any iteration or time $t$. Also, define $f_m(x_m) = \mathbb{E}_{\xi \sim \mathcal{D}_m} f (x_m; \xi)$.

\begin{itemize}
    \item Function $F$ is $L$-smooth.
    \item $\mathbb{E}_{\xi \sim \mathcal{D}_m} || \nabla f(x, \xi) - \nabla f_m(x_m) ||^2 \leq \sigma^2$
    \item $\mathbb{E}_{m \sim U[M]} || \nabla f_m(x_m) - \nabla F(x_m) ||^2 \leq \zeta^2$
    \item $\mathbb{E} || x_m^t - x_c^t  ||^2 \leq \rho^2$ at any time or iteration $t$.    
\end{itemize}

\begin{proof}
We start by using the $L$-smooth property of the function $F$ on the points $x_m^{t+1}$ and $x_m^t$.

\begin{equation}
    \mathbb{E}[F(x_m^{t+1}] \leq \mathbb{E}[F(x_m^t] + \mathbb{E}\langle \nabla F(x_m^t), x_m^{t+1} - x_m^t\rangle + \frac{L}{2} \mathbb{E}||x_m^{t+1} - x_m^t||^2
    \label{nonconvex:L}
\end{equation}

Using the update rule: $x_m^{t+1} = x_m^t - \eta \nabla f(x_m^t, \xi) - \eta \lambda (x_c^t - x_m^t)$, we can write: 

\begin{equation*}
\begin{aligned}
    \mathbb{E}[F(x_m^{t+1})] \leq \mathbb{E}[F(x_m^t)] - \eta \mathbb{E} \langle \nabla F(x_m^t), \nabla f(x_m^t, \xi) \rangle - &\lambda \mathbb{E} \langle \nabla F(x_m^t), x_m^t-x_c^t \rangle \\ 
    + &\frac{L\eta^2}{2} \mathbb{E}||\nabla f(x_m^t, \xi)||^2 + \frac{L\eta^2\lambda^2}{2}  \mathbb{E}||x_m^t-x_c^t||^2 
\end{aligned}
\end{equation*}

With further manipulation of the terms, we obtain the following:

\begin{equation}
\begin{aligned}
    \mathbb{E}[F(x_m^{t+1}] \leq \mathbb{E}[F(x_m^t] - & \left (\eta - L\eta^2 \right ) \mathbb{E} \langle \nabla F(x_m^t), \nabla f(x_m^t, \xi) \rangle  + \frac{L\eta^2}{2} \mathbb{E} || \nabla F(x_m^t) - \nabla f(x_m^t ||^2 \\
    - & \eta \lambda \mathbb{E} \langle \nabla F(x_m^t), x_m^t-x_c^t \rangle + \frac{L\eta^2 \lambda^2}{2}  \mathbb{E}||x_m^t-x_c^t||^2 
\end{aligned}
\label{nonconvex:L_update}
\end{equation}

We will derive upper bounds for each term in \ref{nonconvex:L_update}. We start with $-\mathbb{E} \langle \nabla F(x_m^t), \nabla f(x_m^t, \xi) \rangle$.

\begin{equation}
\begin{aligned}
- \mathbb{E} \langle \nabla F(x_m^t), \nabla f(x_m^t, \xi) \rangle = \frac{1}{2} \mathbb{E} & || \nabla F(x_m^t) - \nabla f(x_m^t, \xi) ||^2 - \frac{1}{2}||\nabla F(x_m^t)||^2 - \frac{1}{2} ||\nabla f(x_m^t, \xi)||^2 \\
\leq & \frac{1}{2} \mathbb{E} || \nabla F(x_m^t) - \nabla f(x_m^t, \xi) ||^2 - \frac{1}{2} \mathbb{E} ||\nabla F(x_m^t)||^2 \\
& \leq \frac{1}{2} \left ( \zeta^2 - \mathbb{E} || \nabla F(x_m^t) ||^2 \right )
\end{aligned}
\label{nonconvex:1}
\end{equation}

First, we derive a bound for the term $\mathbb{E} || \nabla F(x_m^t) - \nabla f(x_m^t ||^2$:
\begin{equation}
\begin{aligned}
\mathbb{E} || \nabla f(x_m^t, \xi) - \nabla F(x_m^t) ||^2 &= \mathbb{E} \ || \nabla f(x_m^t, \xi) - \mathbb{E}_{\xi \sim \mathcal{D}_m } \nabla f (x_m; \xi) - \left ( \nabla F(x_m^t) - \mathbb{E}_{\xi \sim \mathcal{D}_m} \nabla f (x_m; \xi) \right )||^2 \\
\leq & \mathbb{E} || \nabla f(x_m^t, \xi) - \mathbb{E}_{\xi \sim \mathcal{D}_m } \nabla f (x_m; \xi) ||^2 + || \nabla F(x_m^t) - \mathbb{E}_{\xi \sim \mathcal{D}_m} \nabla f (x_m; \xi) ||^2 \\
= & \mathbb{E} || \nabla f(x_m^t, \xi) - \nabla f_m (x_m) ||^2 + || \nabla F(x_m^t) -\nabla f_m (x_m; \xi) ||^2 \leq \sigma^2 + \zeta^2
\end{aligned}
\label{nonconvex:2}
\end{equation}

Lastly, we derive a bound for the term $ - \mathbb{E} \langle \nabla F(x_m^t), x_m^t - x_c^t \rangle$:

\begin{equation}
\begin{aligned}
- \mathbb{E} \langle \nabla F(x_m^t), x_m^t - x_c^t \rangle = \frac{1}{2} \mathbb{E}  ||\nabla F(x_m^t)||^2 + & \frac{1}{2} || x_m^t - x_c^t ||^2 - \frac{1}{2} || \nabla F(x_m^t) + (x_m^t - x_c^t) ||^2 \\
\leq & \frac{1}{2} \mathbb{E} \left ( ||\nabla F(x_m^t)||^2 + || x_m^t - x_c^t ||^2 \right ) \\
\leq & \frac{1}{2} \mathbb{E} ||\nabla F(x_m^t)||^2 + \frac{\rho^2}{2}
\end{aligned}
\label{nonconvex:3}
\end{equation}

By making use of the expressions in \ref{nonconvex:1}, \ref{nonconvex:2} and \ref{nonconvex:3}; we can re-write the inequality in \ref{nonconvex:L_update} as follows:

    

\begin{equation}
\begin{aligned}
\mathbb{E}F(x_m^{t+1})  \leq \mathbb{E}F(x_m^t) + &  \frac{1}{2} \left ( \eta - L\eta^2 \right ) \left ( \zeta^2 - \mathbb{E}||\nabla F(x_m^t) ||^2 \right ) + \frac{L\eta^2}{2} (\zeta^2 + \sigma^2) \\
&+ \frac{\eta\lambda}{2} \mathbb{E} || \nabla F(x_m^t) ||^2  + \frac{\eta\lambda}{2} \rho^2 + \frac{\eta^2\lambda^2L}{r} \rho^2 
\end{aligned}
\end{equation}

\begin{equation}
\left ( \frac{\eta - \lambda - L \eta^2}{2} \right )  \mathbb{E}|| \nabla F(x_m^t) ||^2 \leq F(x_m^t) -  F(x_m^{t+1})+ \left ( \frac{\eta}{2} + L \eta^2 \right ) \zeta^2  + \frac{ L \eta^2 }{2}\sigma^2 + \left ( \frac{\eta \lambda }{2} + \frac{\eta^2 \lambda^2 L}{2} \right ) \rho^2
\end{equation}

We set $\eta \leq \frac{\frac{1}{3} - \lambda}{L}$ so that we satisfy $\frac{\eta (1 - \lambda - L \eta}{2} \geq \frac{\eta}{3}$. As a result, we can write:

\begin{equation}
\mathbb{E}|| \nabla F(x_m^t) ||^2 \leq \frac{3}{\eta} \left ( F(x_m^t) -  F(x_m^{t+1})+ \left ( \frac{\eta}{2} + L \eta^2 \right ) \zeta^2  + \frac{ L \eta^2 }{2}\sigma^2 + \left ( \frac{\eta \lambda }{2} + \frac{\eta^2 \lambda^2 L}{2} \right ) \rho^2 \right )
\end{equation}

We then take the average of the sum of the inequality from $t = 1,..., N$ and $m = 1,..., M$; we ultimately get the following inequality where $x^0$ and $x^*$ are the initial model parameters and minimizer of $F$ respectively:

\begin{equation}
\frac{1}{MN}\sum_{m=1}^{M} \sum_{t=1}^{N} \mathbb{E}|| \nabla F(x_m^t) ||^2 \leq \frac{3(F(x^0) -  F(x^*))}{MN\eta} + \frac{3}{2}(\zeta^2 + \lambda\rho^2) + \frac{3}{2} \left ( 2L\zeta^2 + L\sigma^2 + \lambda^2\rho^2L \right ) \eta 
\end{equation}

Note that this convergence rate analysis characterizes the average of the gradient norm square obtained by all local variables on the function $F$.

\end{proof}

\end{theorem}

\subsection{Relating GRAWA and MGRAWA Coefficients to Non-Convex Convergence Analysis}
In this subsection, we directly relate the $\beta_1, \beta_2, ..., \beta_M$ coefficients that appear in the vanilla GRAWA and MGRAWA to one of the conditions imposed in Theorem \ref{appendix:nonconvex}. Notice that, as mentioned before, it is mathematically intractable to find a relation between LGRAWA weights and the convergence rate due to its layer-wise weighted averaging. \\

We keep the first three assumptions of Theorem \ref{appendix:nonconvex}, remove the last one, and add two more. Overall, we have the following set of assumptions:

\begin{itemize}
    \item Function $F$ is $L$-smooth.
    \item $\mathbb{E}_{\xi \sim \mathcal{D}_m} || \nabla f(x, \xi) - \nabla f_m(x_m) ||^2 \leq \sigma^2$
    \item $\mathbb{E}_{m \sim U[M]} || \nabla f_m(x_m) - \nabla F(x_m) ||^2 \leq \zeta^2$
    \item $\mathbb{E} \left [ x_i^{t^T} x_j^t \right ] \leq y^2$ at any time or iteration $t$ for all $ i,j \in [1, M] $.
    \item $\mathbb{E} \left [ \beta_i^t \beta_j^t \right ] \leq \frac{\rho^2 + y^2}{y^2 M^2}$ at any time or iteration $t$ for all $ i,j \in [1, M] $.  
\end{itemize}

The last two assumptions are conditioned on the value of the dot product between any two workers at any iteration and, the value of the multiplication between any two GRAWA/MGRAWA coefficients at any iteration. In the extension, we show that having these two assumptions is equivalent to the last assumption of Theorem \ref{appendix:nonconvex}.

\begin{proof}
We start by writing $x_m^t - x_c^t = x_m^t - \sum_{i=1}^M \beta_i x_i^t$ and expressing $||x_m^t - x_c^t||^2$ as:
\begin{equation*}
\begin{aligned}
    ||x_m^t - x_c^t||^2 = (x_m^t - x_c^t)^T(x_m^t - x_c^t) &= x_m^{t^T} x_m^t -2 \sum_{i=1}^M \beta_i x_m^{t^T} x_i^t + \left (  \sum_{i=1}^M\beta_i x_i^t  \right) ^ T \left (  \sum_{i=1}^M \beta_i x_i^t  \right) \\
    & = x_m^{t^T} x_m^t -2 \sum_{i=1}^M \beta_i x_m^{t^T} x_i^t + \sum_{i=1}^M \sum_{j=1}^M \beta_i \beta_j x_i^{t^T}  x_j^t 
\end{aligned}
\end{equation*}

Taking the expectation of the last expression by also using independence between $\beta_m$ and $x_m$, we arrive at the following inequality:

\begin{equation*}
\begin{aligned}
    \mathbb{E} \left [||x_m^t - x_c^t||^2 \right ] &= \mathbb{E} \left [ x_m^{t^T} x_m^t \right ] -2 \sum_{i=1}^M \beta_i  \mathbb{E} \left [x_m^{t^T} x_i^t \right ] + \sum_{i=1}^M \sum_{j=1}^M \mathbb{E} \left [ \beta_i \beta_j\right ] \mathbb{E} \left [ x_i^{t^T}  x_j^t \right ] \\
    &\leq y^2 - 2 y^2 \sum_{i=1}^M \beta_i + \sum_{i=1}^M \sum_{j=1}^M y^2 \frac{\rho^2 + y^2}{y^2 M^2} \\
    &\leq y^2 - 2 y^2 \sum_{i=1}^M \beta_i + y^2 M^2 \frac{\rho^2 + y^2}{y^2 M^2} 
\end{aligned}
\end{equation*} 

By definition we have $\sum_{i=1}^M \beta_i = 1$, so we can also write:
\begin{equation*}
    \mathbb{E} \left [||x_m^t - x_c^t||^2 \right ] \leq y^2 \left ( 1 - 2 \right) + \rho^2 + y^2 = -y^2 + \rho^2 + y^2
\end{equation*} 

which means that overall we have $\mathbb{E} \left [||x_m^t - x_c^t||^2 \right ] \leq \rho^2$ that is the same as the last assumption specified in Theorem \ref{appendix:nonconvex}.
\end{proof}

\section{Pseudo-Codes of the Competing Algorithms}

Below are the pseudo-codes of two other parameter-sharing-based distributed training algorithms that compete with MGRAWA and LGRAWA. To form the center variable, the EASGD takes the average of the workers in the distributed environment both with respect to space and time whereas, the LSGD chooses the leader worker with the smallest loss value as the center variable.

\begin{minipage}{0.46\textwidth}
\begin{algorithm}[H]
    \caption{EASGD}
    
    \begin{algorithmic}
        \State \textbf{Input}: Pulling force $\lambda$, moving average strength $\rho$, communication period $\tau$, learning rate $\eta$,  loss function $f$
        \State \textbf{Initialize} workers $x_1, x_2, ..., x_M, x_C$ from the same random model, worker-exclusive data shards $\Psi_1, \Psi_2, ..., \Psi_M$ and iteration counters for workers $t_1=t_2= ... =t_M=0$ 
        \State At each worker $m$ \textbf{do} 
            \While{not converged}
            \State Draw a random batch $\xi_m \in \Psi_m$
            \State $x_{m} \leftarrow x_{m} - \eta \nabla f(x_m; \xi_m) $  
            \State $t_m \leftarrow t_m + 1$
            
            \If{$M\tau$ divides $\sum_{m=1}^{M} t_m$}
            \State $x_a = \sum_{m=1}^{M} \beta_{m} x_m$   
            \State $x_C \leftarrow (1-\rho) x_a + \rho x_a$ 
            \State $x_m \leftarrow (1-\lambda) x_m + \lambda x_C$ 
            \EndIf
            \EndWhile

    \end{algorithmic}
\end{algorithm}
\end{minipage}
\hfill
\begin{minipage}{0.46\textwidth}
\begin{algorithm}[H]
    \caption{LSGD}
    
    \begin{algorithmic}
        \State \textbf{Input}: Pulling force $\lambda$, proximity force strength $\mu$, communication period $\tau$, learning rate $\eta$,  loss function $f$
    \State \textbf{Initialize} workers $x_1, x_2, ..., x_M, x_C$ from the same random model, worker-exclusive data shards $\Psi_1, \Psi_2, ..., \Psi_M$ and iteration counters for workers $t_1=t_2= ... =t_M=0$ 
        \State At each worker $m$ \textbf{do} 
        \While{not converged}
        \State Draw a random batch $\xi_m \in \Psi_m$
        \State $x_{m} \leftarrow x_{m} - \eta \nabla f(x_m; \xi_m) $  
        \State $x_m \leftarrow (1-\frac{\mu}{\tau}) x_m + \frac{\mu}{\tau} x_C$ (Prox.)
        \State $t_m \leftarrow t_m + 1$
        
        \If{$M\tau$ divides $\sum_{m=1}^{M} t_m$}
        \State $x_C \leftarrow \text{argmin}_{x_i} f(x_i, \xi_i)$ 
        \State $x_m \leftarrow (1-\lambda) x_m + \lambda x_C$ 
        \EndIf
        \EndWhile
    \end{algorithmic}
\end{algorithm}
\end{minipage}

We also share the pseudo-codes of DP+SGD and DP+SAM configurations. The execution of the SAM optimizer in the parallel environment is not as straightforward. Per the official implementation suggestion, the ascend step is not synchronized across the distributed processes but the descend step is synchronized.

\begin{minipage}{0.46\textwidth}
\begin{algorithm}[H]
    \caption{DP+SGD}
    
    \begin{algorithmic}
        \State \textbf{Input}: learning rate $\eta$, loss function $f$
        \State \textbf{Initialize} workers $x_1, x_2, ..., x_M, x_C$ from the same random model and worker-exclusive data shards $\Psi_1, \Psi_2, ..., \Psi_M$
        \State At each worker $m$ \textbf{do} 
        \While{not converged}
        \State Draw a random batch $\xi_m \in \Psi_m$
        \State $g_m \leftarrow \nabla f(x_m; \xi_m) $  
        \State \textbf{Initiate} distributed communication
        \State \hskip 1.5em $g = \frac{1}{M}\sum_{m=1}^{M} g_m$ 
        \State $x_{m} \leftarrow x_{m} - \eta g $ 
        \EndWhile
    \end{algorithmic}
\end{algorithm}
\end{minipage}
\hfill
\begin{minipage}{0.46\textwidth}
\begin{algorithm}[H]
    \caption{DP+SAM}
    
    \begin{algorithmic}
        \State \textbf{Input}: Pulling force $\lambda$, moving average strength $\rho$, communication period $\tau$, learning rate $\eta$,  loss function $f$
    \State \textbf{Initialize} workers $x_1, x_2, ..., x_M, x_C$ from the same random model and worker-exclusive data shards $\Psi_1, \Psi_2, ..., \Psi_M$
        \State At each worker $m$ \textbf{do} 
        \While{not converged}
        \State Draw a random batch $\xi_m \in \Psi_m$
        \State $x_{m,adv} \leftarrow x_{m} + \rho \frac{\nabla f(x_m; \xi_m)}{||\nabla f(x_m; \xi_m)||} $ 
        
        \State $g_{m,adv} \leftarrow \nabla f(x_{m,adv}; \xi_m) $  
        \State \textbf{Initiate} distributed communication
        \State \hskip 1.5em $g = \frac{1}{M}\sum_{m=1}^{M} g_{m,adv}$ 
        \State $x_{m} \leftarrow x_{m} - \eta g $ 
        \EndWhile
    \end{algorithmic}
\end{algorithm}
\end{minipage}

\section{Local Optimization Step Aware GRAWA: Local GRAWA}
We also propose versions of MGRAWA and LGRAWA that use the gradient information from the local optimizer rather than accumulating the gradient on a separate batch that is exclusively used for this purpose. We refer to these variants Local-MGRAWA and Local-LGRAWA respectively. Particularly, the gradient accumulation step is not present and instead, the gradient norms are calculated and kept track of from the batches encountered by the local optimizer such as SGD. \\

At the implementation level, we wrap the local optimizer to access the gradient information from its state and use it to calculate gradient norms for each batch suitable to MGRAWA or LGRAWA, depending on the chosen type. Note that the model-level and layer-level gradient norm aware flat minima-seeking mechanisms are still present for MGRAWA and LGRAWA. The difference comes from directly using local optimizer gradients rather than dedicating a phase to selecting another random batch and calculating gradients with it. Until the communication phase, each worker keeps track of the running estimate of the gradient norm encountered by its data shard and then the estimate is used for assigning weights to the worker's model or layers depending on MGRAWA and LGRAWA selection. The pseudo-code can be found below. \\

\begin{minipage}{1\textwidth}
\centering
\begin{algorithm}[H]
    \caption{Local MGRAWA/LGRAWA}

    \begin{algorithmic}
        \State \textbf{Input}: Pulling force $\lambda$, grad. norm momentum $\gamma$, communication period $\tau$, learning rate $\eta$,  loss function $f$
    \State \textbf{Initialize} workers $x_1, x_2, ..., x_M, x_C$ from the same random model and worker-exclusive data shards $\Psi_1, \Psi_2, ..., \Psi_M$
        \State At each worker $m$ \textbf{do} 
        \State $g_{mvg}=concat\left [0, 0, ..., 0 \right ]$
        \While{not converged}
        \State Draw a random batch $\xi_m \in \Psi_m$ and form $g_{cur} = concat\left [||g_1||, ||g_2||, ..., ||g_K|| \right ]$
        \State $t_m \leftarrow t_m + 1$ 
        \State $g_{mvg} \leftarrow \gamma g_{mvg} + (1- \gamma) g_{cur}$ 
        \State $g_{est} \leftarrow \gamma g_{mvg} / (1 - \gamma^{t_m})$ 
        \State $x_{m} \leftarrow x_{m} - \eta \nabla f(x_m; \xi_m) $  
        \State $x_m \leftarrow (1-\frac{\mu}{\tau}) x_m + \frac{\mu}{\tau} x_C$ (Prox.)
        
        \If{$M\tau$ divides $\sum_{m=1}^{M} t_m$}
        \State form $x_C$ using the running gradient norm estimate $g_{est}$ based on MGRAWA/LGRAWA policy
        \State $x_m \leftarrow (1-\lambda) x_m + \lambda x_C$ 
        \State $g_{mvg} \leftarrow concat\left [0, 0, ..., 0 \right ]$
        \State $t_m \leftarrow 0$
        \EndIf
        \EndWhile
    \end{algorithmic}
\end{algorithm}
\end{minipage}

\end{document}